\documentclass[lettersize,journal]{IEEEtran}
\usepackage{amsmath}
\usepackage{amsfonts}
\usepackage{comment}
\usepackage{algorithmic}
\usepackage{algorithm}
\usepackage{array}
\usepackage{textcomp}
\usepackage{stfloats}
\usepackage{url}
\usepackage{verbatim}
\usepackage{graphicx}
\usepackage{cite}
\usepackage{pifont}
\usepackage{float}
\usepackage{bbm}
\usepackage{bm}
\usepackage{media9}
\usepackage{hyperref}
\newcommand{\cmark}{\ding{51}} %
\newcommand{\xmark}{\ding{55}} %
\usepackage{multirow}
\usepackage{booktabs}
\usepackage{color}

\begin{document}

\title{Sounding Video Generator: A Unified Framework for Text-guided Sounding Video Generation}

\author{Jiawei Liu, Weining Wang, Sihan Chen, Xinxin Zhu, Jing Liu$^{*}$

\thanks{* Corresponding author.}
\thanks{Jiawei Liu, Weining Wang, Sihan Chen, Xinxin Zhu and Jing Liu are with The Laboratory of Cognition and Decision Intelligence for Complex Systems, Institute of Automation, Chinese Academy of Sciences (e-mail: liujiawei2020@ia.ac.cn, weining.wang@nlpr.ia.ac.cn, chensihan2019@ia.ac.cn, xinxin.zhu@nlpr.ia.ac.cn, jliu@nlpr.ia.ac.cn).}
\thanks{Jiawei Liu, Sihan Chen and Jing Liu are also with School of Artificial Intelligence, University of Chinese Academy of Sciences.}
\thanks{Manuscript created October 31, 2022, revised February 13, 2023.}%
}



\maketitle

\begin{abstract}
 As a combination of visual and audio signals, video is inherently multi-modal.
 However, existing video generation methods are primarily intended for the synthesis of visual frames, whereas audio signals in realistic videos are disregarded. 
 In this work, we concentrate on a rarely investigated problem of text-guided sounding video generation and propose the \textbf{S}ounding \textbf{V}ideo \textbf{G}enerator (SVG), a unified framework for generating realistic videos along with audio signals.  
 Specifically, we present the SVG-VQGAN to transform visual frames and audio mel-spectrograms into discrete tokens.
 SVG-VQGAN applies a novel hybrid contrastive learning method to model inter-modal and intra-modal consistency and improve the quantized representations.
 A cross-modal attention module is employed to extract associated features of visual frames and audio signals for contrastive learning.
  Then, a Transformer-based decoder is used to model associations between texts, visual frames, and audio signals at token level for auto-regressive sounding video generation. 
  AudioSet-Cap, a human annotated text-video-audio paired dataset, is produced for training SVG. 
  Experimental results demonstrate the superiority of our method when compared with existing 
  text-to-video generation methods as well as audio generation methods on Kinetics and VAS datasets. 
\end{abstract}


\section{Introduction}
\begin{figure}
  \centering
  \includegraphics[width=\linewidth]{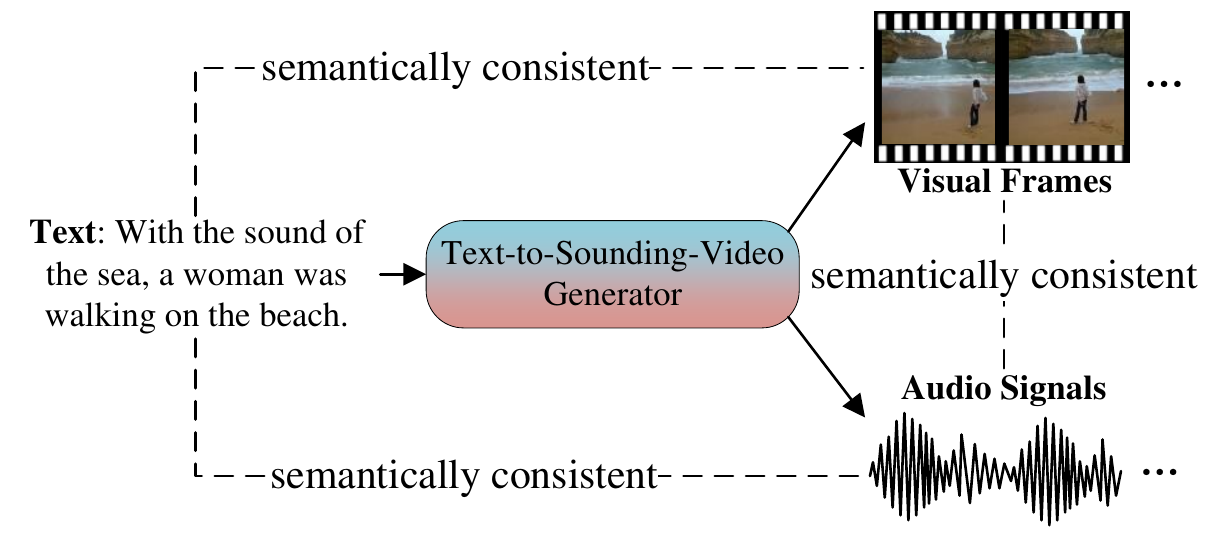}
  \caption{Illustration of the proposed text-to-sounding-video generation task. 
}
  \label{fig:task}
  \vspace{-3mm}
\end{figure}

\IEEEPARstart{V}{ideo} generation \cite{vgan,mocogan,mocogan-hd} has attracted a lot of attention from both academia and industry, since it has the ability to generate videos without copyright issues for media makers and aid in data augmentation for deep learning models.
Text-to-video generation \cite{t2v,godiva,nuwa}, in particular, which synthesises videos with natural language as a condition, has improved controllability and is becoming a popular research subject. 
Current text-to-video generation approaches mainly concentrate on visual frame generation.
However, video is actually a type of multi-modal data that includes both visual and audio components.
Videos with background audio signals, i.e., sounding videos, include more comprehensive information and are beneficial to video understanding for both humans and machines \cite{videosum}.
For example, it is hard to determine whether a person in a video is singing or speaking without audio.
Therefore, as shown in Fig. \ref{fig:task}, we propose a novel task of \textbf{T}ext-to-\textbf{S}ounding-\textbf{V}ideo (T2SV) generation that synthesizes high fidelity sounding videos semantically consistent with the guided textual descriptions. 

Three factors are essential for successful T2SV generation:
\textbf{(1)} How to model cross-modal associations for better video representation? In such a multi-modal data as video, cross-modal associations occur naturally and can enable us to obtain more comprehensive and semantically distinct video representations. For instance, using audio information can help identify visually similar objects, such as horses and donkeys.

\textbf{(2)} It is difficult to generate visual and audio content that is consistent with the guided text while ensuring the correlation and timing alignment of visual frames and audio signals. Tri-modal semantic consistency must be modeled during the generation process. 
\textbf{(3)} There is no paired text-video-audio dataset that contains textual descriptions for both visual and audio content. 
Previous text-video paired datasets \cite{webvid,howto} concentrate mostly on the visual content and omit the descriptions of audio, whereas the T2SV task needs semantic congruence between audio and text.

\begin{figure*}
  \centering
  \includegraphics[width=\linewidth]{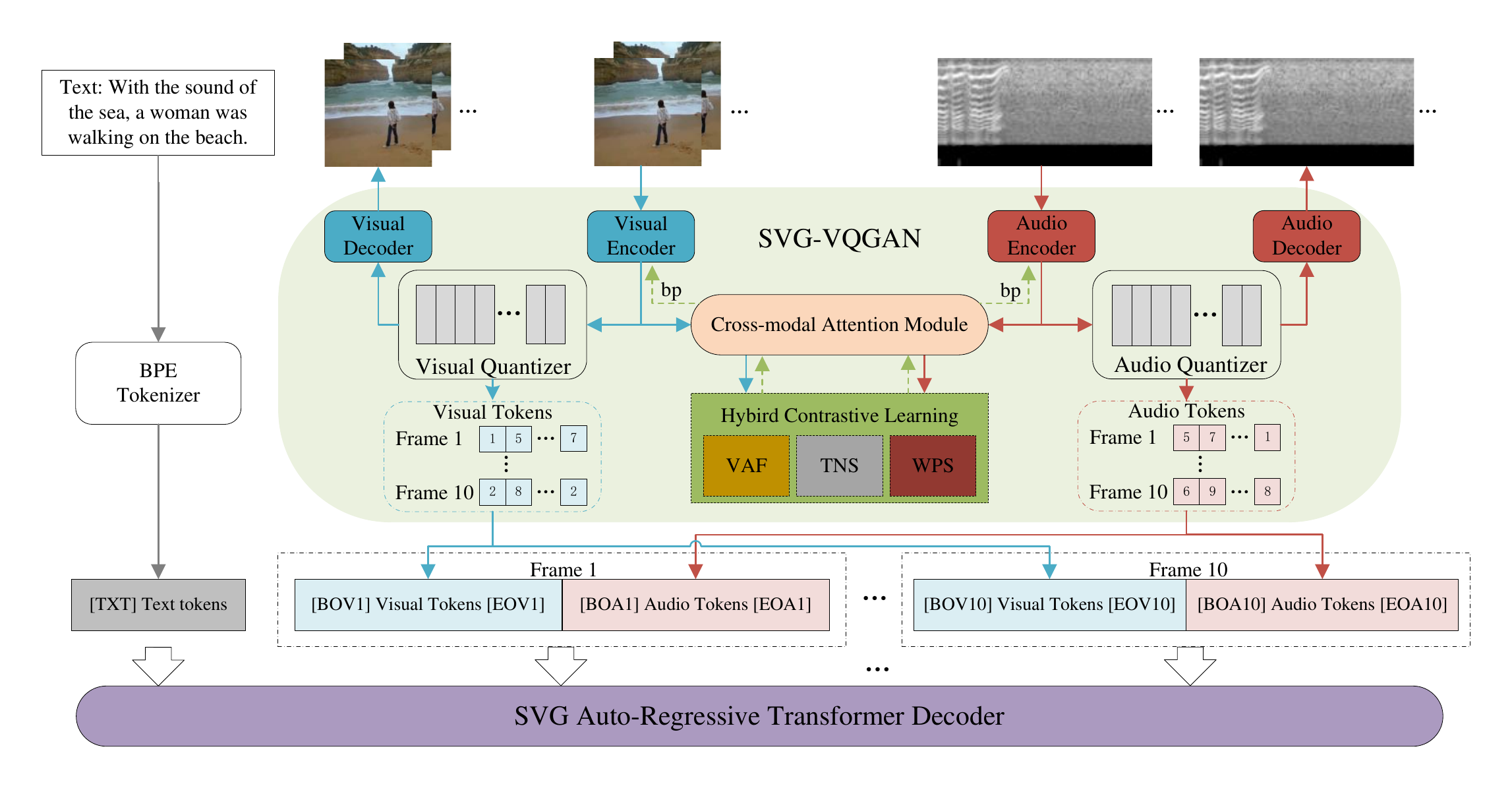}
  \setlength{\abovecaptionskip}{-5mm}
  \caption{Overview of the proposed SVG framework.
  Text is tokenized by BPE tokenizer. Visual frames and audio spectrograms are tokenized by the proposed SVG-VQGAN with Cross-modal Attention Module and Hybrid Contrastive Learning modeling visual-audio associations. The dotted green line indicates the back propagation (bp) of contrastive loss. The Visual-Audio-similarity-based Filter (VAF), Text-guided Negative samples Selection (TNS) and Window-based Positive samples Selection (WPS) strategies are used to refine the positive and negative samples in contrastive learning. Then an auto-regressive Transformer decoder is used to generate visual and audio content frame by frame and token by token. 
}
  \label{fig:framework}
\end{figure*}

To address the above issues or challenges, we propose a novel \textbf{S}ounding \textbf{V}ideo \textbf{G}enerator (SVG). 
As shown in Fig. \ref{fig:framework}, 
SVG consists of two stages: quantized encoding and discrete token generation.
Firstly, the visual frames and audio spectrograms are independently quantized into discrete tokens using a two-stream SVG-\textbf{V}ector-\textbf{Q}uantized GAN (SVG-VQGAN). 
To obtain better quantized representations, we propose a hybrid contrastive learning method,
in which inter-modal contrastive loss is adopted to model cross-modal associations,
and intra-modal contrastive loss is employed as a regularization to prevent the extracted features from straying away from the original modality.
We select positive and negative samples from the same and different video clips separately.
To further refine the selection process, we propose three strategies: visual-audio-similarity-based filter, text-guided negative samples selection and window-based positive samples selection. 
Notably, some visual entities, such as the sky background, have no corresponding audio counterparts, and the same holds true for audio.
Thus, a cross-modal attention module is proposed to build local alignment for visual and audio content, and obtain the global features for hybrid contrastive learning.
Then, at the second stage, an auto-regressive Transformer decoder is adopted to model semantic consistency between text descriptions, visual frames, and audio signals triples at the token level.
To {take both visual-to-audio and audio-to-visual attention into account}, we suggest a modality alternate sequence format where visual tokens and audio tokens are concatenated in each frame and then cascaded frame by frame.

To compensate for the lack of appropriate datasets, AudioSet-Cap, a human annotated text-video-audio paired dataset, is produced for training SVG.
AudioSet-Cap is a large-scale dataset that contains audio-rich videos from AudioSet\cite{audioset}. 
Every video in AudioSet-Cap is annotated by a human annotator with a caption describing both the visual and audio content, whereas previous text-video paired datasets only describe the visual content. Consequently, AudioSet-Cap is a more appropriate dataset for the T2SV task.

The main contributions of this work are four-folds:
\begin{itemize}
\item This is the first work to focus on a novel task of text to sounding video generation using a unified framework.
\item We propose a novel SVG-VQGAN, where a cross-modal attention module is introduced to build local semantic correspondence and hybrid contrastive learning is proposed to model inter-modal and intra-modal consistency.
\item A human annotated dataset, with descriptions for both visual and audio content,  is produced for T2SV generation. 
\item Experimental results demonstrate that SVG achieves excellent performance on T2SV, text-to-video, and open-domain audio generation tasks with the proposed SVG-VQGAN and modality alternate sequence format.
\end{itemize}

\section{Related Works}
Recent years witnessed significant progress in the understanding and generation tasks of visual \cite{cnn,li2021ctnet,dalle} and audio \cite{hifigan,audeo} content.
The following multimodal works \cite{clip,videosum,globallocal} place great emphasis on multimodal joint understanding.
This paper proposes a novel task for the joint generation of visual and audio content, i.e., T2SV. 
In this section, we briefly review related works for video generation and audio generation. 
Notably, the primary distinction between prior works and ours is that we model visual-audio association for video representation and generate video with background audio signals using a unified model.
\subsection{Video Generation}
Previous video generation works can be divided into one-stage methods based on GAN \cite{gan} and two-stage methods based on \textbf{V}ector \textbf{Q}uantized VAE (VQVAE) \cite{vqvae} and Transformer \cite{Transformer}. 

GAN-based one-stage methods have achieved excellent performances for video-to-video generation on in-domain datasets \cite{bair, ucf101}, by separating spatio-temporal generation \cite{vgan,tgan} or disentangling motion and content \cite{mocogan,mocogan-hd,DMEE}, etc. 
As for text-to-video generation, RNN is used  to extract text features and generate gist for video generator constructed from 3D convolutional GAN \cite{t2v}. 
TF-GAN \cite{tfgan} proposes a text-conditioning scheme on frame-scale and video-scale, which improves text-video associations. 
However, those GAN-based methods are hard to extend to open-domain scenarios, limited by the training stability and robustness of GANs.

VQVAE \cite{vqvae, vqvae2} and Transformer \cite{Transformer} based auto-regressive generation models have been popular for image and video generation task. Models like DALLE \cite{dalle} and Cogview \cite{cogview} have achieved significant progress on open-domain text-to-image generation, where discrete visual tokens enable efficient and large-scale training of Transformers. LVT \cite{lvt} and GODIVA \cite{godiva} use 2D frame VQVAEs to transform visual frames into discrete tokens and VideoGPT \cite{videogpt} then proposes a 3D version. 
N{\"U}WA \cite{nuwa} uses frame VQGAN \cite{taming} taking advantage of GAN to improve the generation fidelity. 
Different from those visual-only methods, audio information is further considered in our proposed SVG-VQGAN.
{ CogVideo \cite{cogvideo} generates a image by a pretrained text-to-image generation model first and then generates subsequent frames.}
We adopt the Transformer in Cogview \cite{cogview}, and modality alternate sequence format is introduced for generating video with corresponding audio signals.

{ Existing interactive multi-modal physical simulators, such as TDW \cite{threedworld}, could simulate high-fidelity visual and audio content, which could also be used for sounding video generation.
However, the variety of simulated videos is limited by the Unity3D Engine, while we focus on the open-domain video generation guided by the text condition.}
\subsection{Audio Generation}
Most of previous audio generation works focus on a specific domain. 
FastSpeech\cite{fastspeech} uses non-auto-regressive Transformer with teacher-student framework to cover the task of text-to-speech generation. 
{ Vis \cite{vis} builds a model based on CNN \cite{cnn} and LSTM \cite{lstm} to synthesize plausible impact sounds from silent videos.
Another popular audio generation task is music synthesis \cite{zhao2018sound,foley}.
For instance, Audeo\cite{audeo} covers the task of generating piano music for a silent performance video, where visual frames are translated into raw mechanical musical symbolic to synthesize temporal correlated music.
DDT \cite{ddt} takes visual motions into account and could perform audio-visual source separation of different instruments robustly.}
CMT\cite{cmt} further focuses on video background music generation and establishes the rhythmic relations between video and background music, with a controllable music Transformer. 

The most similar work to ours is SpecVQGAN \cite{specvqgan}, which addresses the task of open-domain audio generation. Different from SpecVQGAN \cite{specvqgan}, we generate sounding videos given a text description, while SpecVQGAN \cite{specvqgan} takes audio class names and video features as input and only generates audio signals. 
SpecVQGAN \cite{specvqgan} discretizes mel-spectrograms and uses a MelGAN \cite{melgan} vocoder to decode audio from mel-spectrograms. 
In this work, visual information is further utilized for audio representation by hybrid contrastive loss and a HifiGAN \cite{hifigan} trained on large-scale dataset is adopted to reconstruct the raw audio signals.  

\section{Method}

We address the task of \textbf{T}ext-to-\textbf{S}ounding-\textbf{V}ideo (T2SV) generation for the first time. 
Formally, let $\textbf{v}=\{v_1,v_2,...,v_L\}$ denotes the $L$ frames of a video and $\textbf{a}$ denotes the audio signal, where $v_i\in \mathbb{R}^{C\times H\times W}$ denotes the $i$-th frame and $C,H,W$ are the channels, height and width of visual frames, respectively. 
The T2SV task can be expressed as: given an input text $\textbf{t}$, a generative model $\textbf{G}$ is required to synthesize visual frames \textbf{v} and background audio signals \textbf{a} by maximizing the posterior probability distribution: 
\begin{equation}
    \textbf{v}, \textbf{a}=\textbf{G}(\textbf{t})=\mathop{\arg\max}\limits_{\textbf{v},\textbf{a}} P(\textbf{v},\textbf{a}|\textbf{t}).
\end{equation}
In this work, we propose SVG, a novel unified framework for T2SV generation, as shown in Fig. \ref{fig:framework}. The mel-spectrogram is extracted from the audio $\textbf{a}$ as $\textbf{m}\in \mathbb{R}^{F\times T}$. 
{ To model temporal correlations, the video clip is uniformly cropped into $L$ sub-clips with $1$ frame in each sub-clip, and $\textbf{m}$ is cropped into $L$ audio frames as $\textbf{m}=\{m_1,m_2,...,m_L\}$, where $m_i\in\mathbb{R}^{F\times\frac{T}{L}}$ denotes the $i$-th audio frame.  
During the training process, we first train the SVG-VQGAN to quantize the visual frames and audio mel-spectrograms into discrete tokens as a reconstruction task.
Then a Transformer decoder is trained with text-visual-audio tokens as input and output in an auto-regressive way, i.e., left-to-right prediction. 
The inference process consists of three parts: 1) The text token is input into the auto-regressive Transformer decoder to generate the matching visual tokens and audio tokens; 2) The generated visual tokens and audio tokens are restored to the visual frames and audio mel-spectrogram through the decoders of SVG-VQGAN; 3) The audio mel-spectrogram is restored to the audio signal through the pre-trained HiFiGAN \cite{hifigan} and combined with the generated visual frames to form the generated sounding video.}

\subsection{SVG-VQGAN}
\label{sec:SVG-VQGAN}
\paragraph{\textbf{Two-Stream VQGAN}}
Two separate 2D VQGANs \cite{taming} for visual frames and audio mel-spectrograms are used as the baseline of SVG-VQGAN. 
First, the visual frames and audio mel-spectrograms of the $i$-th to $j$-th frame randomly sampled from a video clip are encoded into visual features $z^v_{i:j}=\{z^v_i,...,z^v_j\}$ and audio features $z^a_{i:j}$:
\begin{equation}
\begin{aligned}
    z^v_k&=E_v(v_k)\in \mathbb{R}^{d_v\times h\times w}, k=i,...,j,\\
    & h=\frac{H}{ds_v}, w=\frac{W}{ds_v}\\
    z^a_{i:j}&=E_a([m_i,...,m_j])\in\mathbb{R}^{d_a\times f\times [(j-i)\times t]},\\
    &f=\frac{F}{ds_a}, t=\frac{T}{L\times ds_a}
\end{aligned}
\label{eq:encode}
\end{equation}
where $E_v$ and $E_a$ denote the encoders of visual frames and audio mel-spectrograms, $ds_v$ and $ds_a$ denote the downsampling rate of $E_v$ and $E_a$, $d_v$ and $d_a$ are the dimension of encoded visual and audio features. 
{ Visual frames are encoded separately. And all the audio frames are concatenated and encoded together because mel-spectrograms are continuous in the time dimension.
$z^v_{i:j}$ and $z^a_{i:j}$ are further mapped into their nearest entries in the visual codebook and audio codebook, respectively. 
Shared quantizer is not adopted for higher reconstruction upper limit.}
In this way, quantized video representations $\hat{z}^v_{i:j}$
and quantized audio representations $\hat{z}^a_{i:j}$
are obtained.
Then, visual frames and audio mel-spectrograms could be reconstructed by VQGAN decoders as $\hat{v}_{i:j}$ and $\hat{m}_{i:j}$. 
The training losses of visual VQGAN and audio VQGAN are formulated as:
\begin{small}
\begin{equation}
\begin{aligned}
    \mathcal{L}^v_{VQGAN}&=\underbrace{\|v_{i:j}-\hat{v}_{i:j}\|_2^2}_{\text{{Reconstruction Loss}}}+\underbrace{\|CNN(v_{i:j})-CNN(\hat{v}_{i:j})\|^2_2}_{\text{Perceptual Loss}}\\
    &+\underbrace{\|z^v_{i:j}-sg(\hat{z}^v_{i:j})\|^2_2+\beta\|sg(z^v_{i:j})-\hat{z}^v_{i:j}\|^2_2}_{\text{Codebook Loss}}\\
    &+\underbrace{\log D^v(v_{i:j})+\log (1-D^v(\hat{v}_{i:j}))}_{\text{Adversarial Loss}}
\end{aligned}
\label{eq:vqloss_video}
\end{equation}
\begin{equation}
\begin{aligned}
    \mathcal{L}^a_{VQGAN}&=\underbrace{\|m_{i:j}-\hat{m}_{i:j}\|_2^2}_{\text{{ Reconstruction Loss}}}+\underbrace{\|CNN(m_{i:j})-CNN(\hat{m}_{i:j})\|^2_2}_{\text{Perceptual Loss}}\\
    &+\underbrace{\|z^a_{i:j}-sg(\hat{z}^a_{i:j})\|^2_2+\beta\|sg(z^a_{i:j})-\hat{z}^a_{i:j}\|^2_2}_{\text{Codebook Loss}}\\
    &+\underbrace{\log D^a(m_{i:j})+\log (1-D^a(\hat{m}_{i:j}))}_{\text{Adversarial Loss}}
\end{aligned}
\label{eq:vqloss_audio}
\end{equation}
\end{small}
where $\beta$ is the weight in codebook loss \cite{vqvae}, $sg$ is the stop-gradient operation, $D^v$ and $D^a$ are patch-based discriminators, the CNN in perceptual loss is VGG-16 \cite{vgg} network pretrained on ImageNet \cite{imagenet} as in VQGAN \cite{taming}.

\paragraph{\textbf{Cross-modal Attention Module}}
For modeling cross-modal associations at the encoding stage, the key idea is that visual frames and audio signals should have semantic correspondence in time, as in previous video-audio self-supervised Learning studies \cite{xdc, slowfast,cmac,globallocal}. 
{ However, it is worth noting that not all visual entities have their associated sound counterparts, e.g., the visual entity `sky' has no associated sound counterparts as it cannot make a sound, and the same is true for audio.}  
Thus, 
a \textbf{C}ross-modal \textbf{A}ttention \textbf{M}odule (CAM) is further proposed to model local semantic associations between visual frames and audio signals.

\begin{figure}
    \centering
    \includegraphics[width=\linewidth]{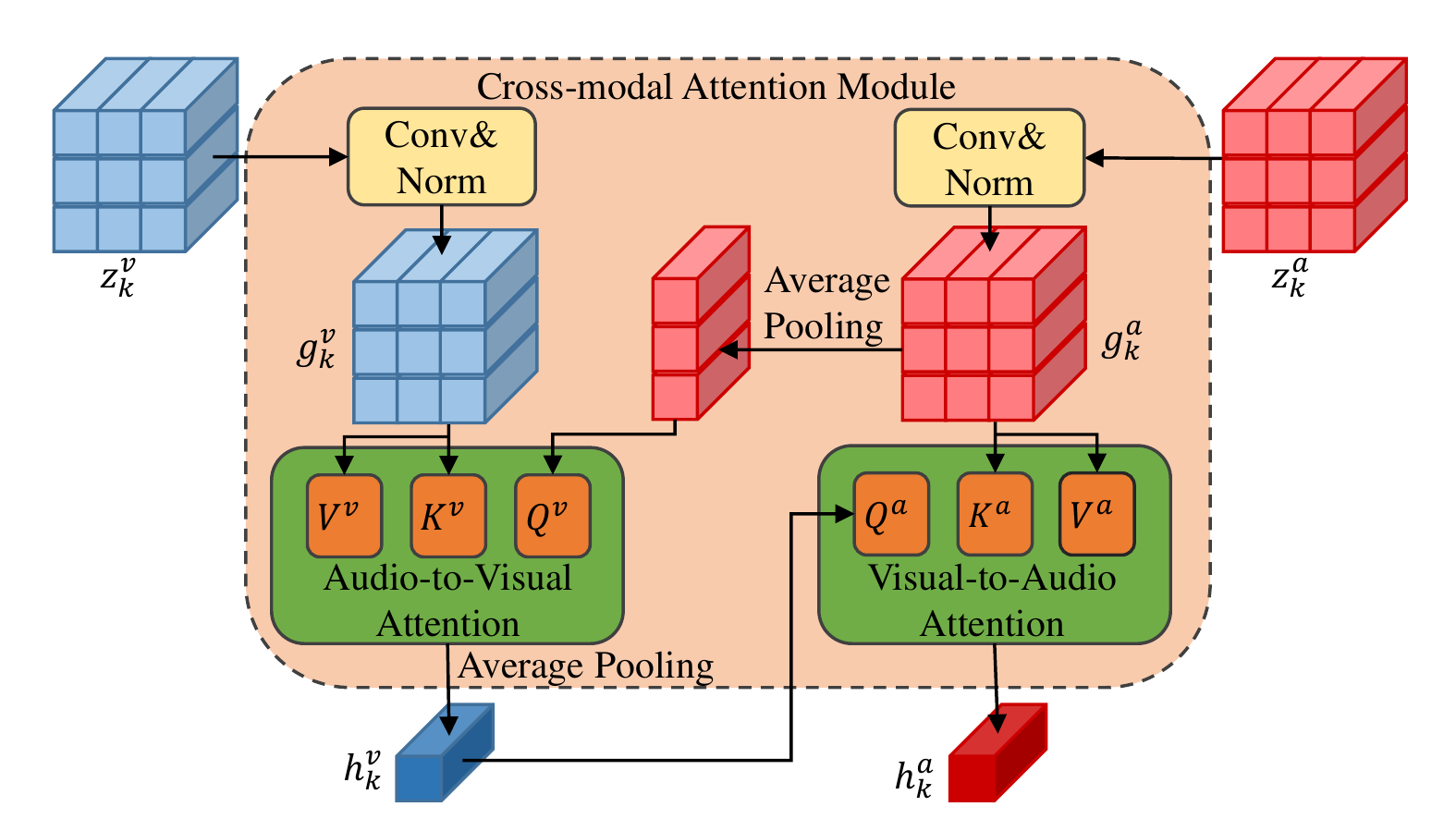}
    \setlength{\abovecaptionskip}{-3mm}
    \caption{Details of the Cross-modal Attention Module.}
    \label{fig:cam}
    \vspace{-2mm}
\end{figure}

As shown in Fig. \ref{fig:cam}, the encoded features of the $k$-th visual frame $z^v_k$ and audio frame $z^a_k$ are first mapped into a common space with several convolutional layers and group normalization \cite{groupnorm}, as $g^v_k$ and $g^a_k$. 
Then $g^a_k$ is averaged in the time dimension, { as visual and audio content may not strictly correspond in time.}
The average pooled $g^a_k$ is taken as the query of audio-to-visual attention, with $g^v_k$ as key and value. The visual features obtained by audio-to-visual attention is further averaged to get a global visual feature of the $k$-th frame as $h^v_k$. Since the audio-associated global visual feature $h^v_k$ has been obtained, we can use it to extract the visual-associated global audio feature of the $k$-th frame by taking $h^v_k$ as the query and $g^a_k$ as the key and value in visual-to-audio attention. 
Both of the audio-to-visual attention and visual-to-audio attention are calculated as in Eq. \ref{eq:att} with input query, key and value as $e^q,e^k,e^v$.
\begin{equation}
\begin{aligned}
  q=Q(e^q),\quad k=K(e^k),\quad v=V(e^v)\\
  h=softmax(\frac{q^T \cdot k}{\sqrt{d^{in}}}) \cdot v
\end{aligned}
\label{eq:att}
\end{equation}
where $Q$, $K$ and $V$ are linear layers, $d^{in}$ is the dimension of $e^q,e^k,e^v$ and $h$ is the output feature. 

\paragraph{\textbf{{Hybrid} Contrastive Learning}}
In the training phase of SVG-VQGAN, we incorporate the objective of modeling the associations between visual frames and audio signals by contrastive learning. 
The visual and audio features in the same video clip should be more consistent than those in video clips with distinct semantics.
The inter-modal contrastive loss is adopted based on this hypothesis, in which $h^{v}$ and $h^a$ from the same video clips are taken as positive samples and those in different video clips are taken as negative samples.
To avoid the extracted features from straying significantly from the original modality, the intra-modal contrastive loss is utilized as a regularization. 
This method is referred to as \textbf{H}ybrid \textbf{C}ontrastive \textbf{L}earning (HCL).
\begin{figure}
    \centering
    \includegraphics[width=\linewidth]{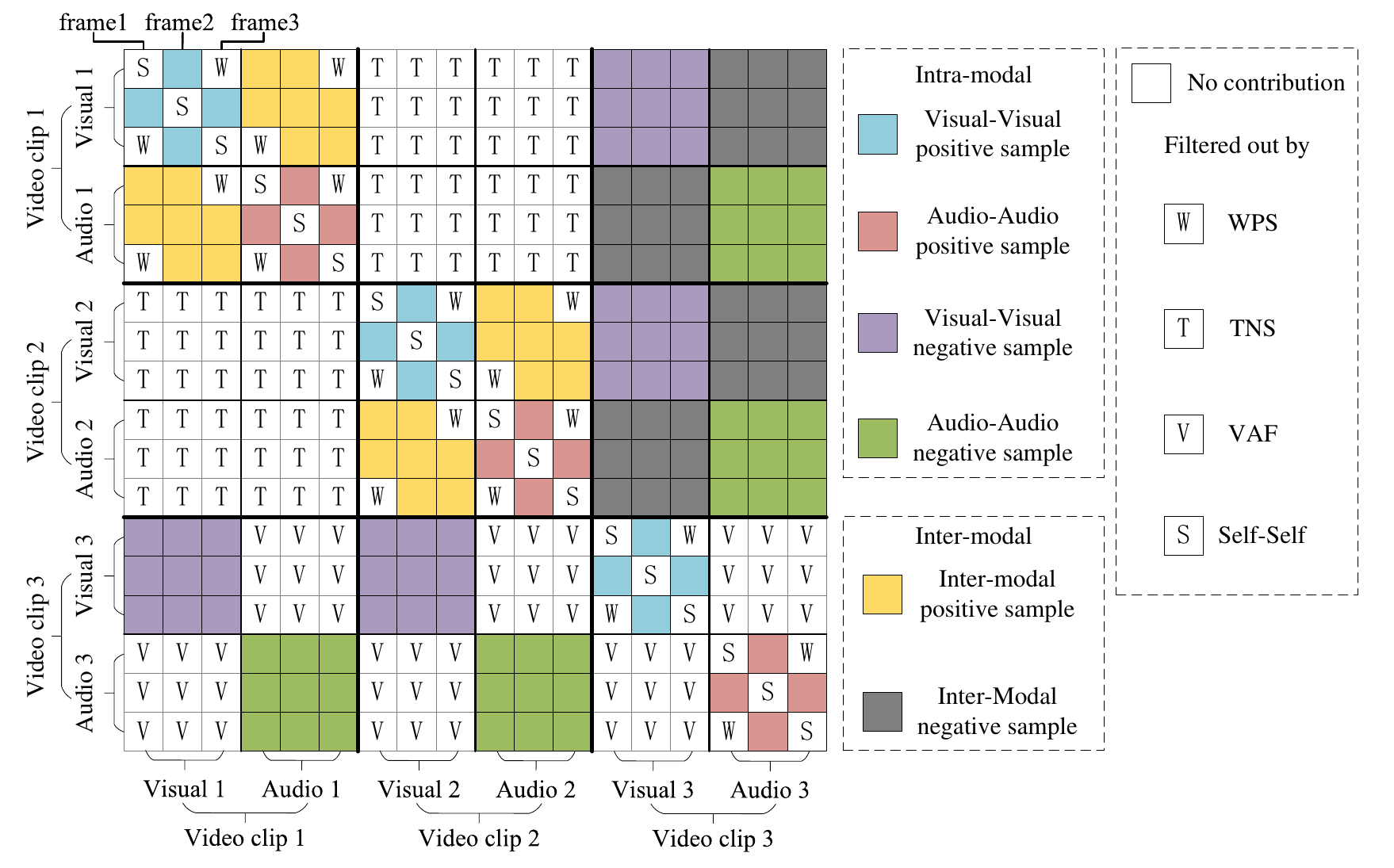}
    \setlength{\abovecaptionskip}{-5mm}
    \caption{Example of positive and negative sample selection with video length of 3, batch size of 3 and WPS window size of 2. Video clip 1 2 have semantically similar texts and will not serve as negative samples of the other. The VAF filters out video clip 3 as its visual and audio content are weakly connected and will not serve as positive samples for inter-modal contrastive learning.
    {The dot product of a feature and itself (self-self) will not be involved in the calculation of contrastive loss as in \cite{supcon}.}}
    \label{fig:selection}
    \vspace{-4mm}
\end{figure}
Two kinds of contrastive losses, i.e. modality split and modality gathered are exploited. 
The modality split version calculates contrastive loss in visual-visual, audio-audio and visual-audio separately, while the modality gathered version takes all visual and audio features equally. 
The loss of modality split HCL ($\mathcal{L}^{MS}_{HCL}$) and modality gathered HCL ($\mathcal{L}^{MG}_{HCL}$) could be respectively defined by Eq. \ref{eq:HCLMS} and Eq. \ref{eq:HCLMG}:
\begin{equation}
\begin{aligned}
    \mathcal{L}^{MS}_{HCL}&=\underbrace{\mathcal{L}_{CL}(\bm{H}^v,\bm{H}^v)+\mathcal{L}_{CL}(\bm{H}^a,\bm{H}^a)}_{\text{intra-modal}}\\
     &+\underbrace{\mathcal{L}_{CL}(\bm{H}^v,\bm{H}^a)+\mathcal{L}_{CL}(\bm{H}^a,\bm{H}^v)}_{\text{inter-modal}}\\
\end{aligned}
\label{eq:HCLMS}
\end{equation}
\begin{equation}
\mathcal{L}^{MG}_{HCL}=\underbrace{\mathcal{L}_{CL}([\bm{H}^v,\bm{H}^a],[\bm{H}^v,\bm{H}^a])}_{\text{inter-modal and intra-modal}}
\label{eq:HCLMG}
\end{equation}
where $\bm{H}^v$ and $\bm{H}^a$ represent all visual and audio features in a batch and $[\bm{H}^v,\bm{H}^a]$ denotes the concatenation of them. $\mathcal{L}_{CL}$ is the contrastive loss which will be introduced in detail later. 

{ Different from previous contrastive learning methods \cite{globallocal,cmac} used in multi-modal encoding, we introduce hybrid contrastive learning in reconstruction and generation tasks, which requires more accurate positive and negative samples.
Thus, we propose three mechanisms to refine the selection process.}

{ Firstly, it is worth noting that some visual entities and background audio signals are not semantically consistent, e.g., music videos with no person in visual content but with human voice.}
In fact, content tags could be used to retrieval images \cite{li2018deep,li2016weakly,li2020weakly} and provide an intermediary for audio and visual content.
In this case, we propose a \textbf{V}isual-\textbf{A}udio similarity based \textbf{F}ilter (VAF) mechanism to use the powerful CLIP \cite{clip} model for filtering out the inter-modal positive samples with low semantic similarity.
Audio categories are extracted by a pre-trained audio recognition model and are further processed to a sentence with a prompt of `\emph{an image with the sound of \{the audio categories\}}'. 
After that the CLIP cosine similarity between every visual frames in the video and the corresponding audio categories sentence will be calculated and visual-audio pairs with CLIP similarity smaller than a preset threshold will be filtered out. 
Note that those samples will still be the negative samples of other video clips for data diversity. 

{ Secondly, it should also be noted that different video clips may be semantically related.}
Thus, we propose a \textbf{T}ext-guided \textbf{N}egative samples \textbf{S}election (TNS) mechanism. 
Text features are extracted by a pre-trained BERT \cite{bert} and could represent the semantic information of a video clip. Thus, we use them to calculate the semantic similarity between different video clips. 
Video clips with BERT similarity higher than a preset threshold will not be chosen as negative samples.

{ Thirdly, frames in the same video clip may be semantically different, as the video subject is likely to change over time.} 
Since the semantics between adjacent frames are generally the same, we propose a \textbf{W}indow-based \textbf{P}ositive sample \textbf{S}election (WPS) mechanism, which refine the selection of positive sample in a random timing window. 

An example of positive and negative sample selection could be found in Fig. \ref{fig:selection}.
Formally, $h_l$ is defined as the global feature of a visual frame or an audio mel-spectrogram frame, extracted from the cross-modal attention module, and $\bm{H}$ is all of the visual or audio global features in a batch. The contrastive loss could be defined based on supervised contrastive losses \cite{supcon} with VAF, TNS and WPS refining the positive and negative samples. 
Specifically, the positive part $\mathcal{P}(h_l,\bm{H}_2)$ and the negative part $\mathcal{N}(h_l,\bm{H}_2)$ of contrastive loss between a single $h_l$ and another set $\bm{H}_2$ are shown in Eq. \ref{eq:pos} and Eq. \ref{eq:neg}: 
\begin{equation}
    \mathcal{P}(h_l,\bm{H}_2)=\sum_{\substack{h_m\in\bm{H}_2,\\h_m\neq h_l}}\mathbbm{1}_{WPS}(h_l,h_m)\exp(h_l^T\cdot h_m/\tau)
    \label{eq:pos}
\end{equation}
\begin{equation}
\begin{aligned}
    \mathcal{N}(h_l,\bm{H}_2)&=\zeta\sum_{h_n\in\bm{H}_2}\mathbbm{1}_{TNS}(h_l,h_n)\exp(h_l^T\cdot h_n/\tau),\\
    \zeta&=\frac{|\bm{H}_2|}{\sum_{h_n\in\bm{H}_2}\mathbbm{1}_{TNS}(h_l,h_n)}
\end{aligned}
\label{eq:neg}
\end{equation}
where $\tau$ is the temperature coefficient, $\mathbbm{1}_{WPS}(h_l,h_m)$ is a binary indicator for WPS to indicate whether $h_m$ is a positive sample for $h_l$, and $\mathbbm{1}_{TNS}(h_l,h_n)$ is a binary indicator for TNS to indicate whether $h_n$ is a negative sample for $h_l$. 
$\mathbbm{1}_{WPS}(h_l,h_m)$ is set to $1$ when the distance between the frames of $h_l$ and $h_m$ is smaller than a preset window size.
$\mathbbm{1}_{TNS}(h_l,h_n)$ is set to $1$ when the cosine similarity between text features of the video clips is smaller than a threshold. 
The $\zeta$ in $\mathcal{N}(h_l,\bm{H}_2)$ is a coefficient used to balance the loss value caused by the unbalanced number of negative samples, where the numerator $|\bm{H}_2|$ represents the total number of samples in $\bm{H}_2$, and the denominator represents the number of negative samples.
Then the contrastive loss could be defined as:
\begin{equation}
    \mathcal{L}_{CL}^l(h_l, \bm{H}_2)=-\log\frac{\mathcal{P}(h_l,\bm{H}_2)}{\mathcal{N}(h_l,\bm{H}_2)}
    \label{eq:CLl}
\end{equation}

\begin{equation}
    \mathcal{L}_{CL}(\bm{H}_1, \bm{H}_2)=\sum_{h_l\in \bm{H}_1} \frac{\mathbbm{1}_{VAF}(h_l)}{\sum_{h_l\in \bm{H}_1}\mathbbm{1}_{VAF}(h_l)}
    \mathcal{L}_{CL}^l(h_l, \bm{H}_2)
    \label{eq:CL}
\end{equation}
where $\mathbbm{1}_{VAF}(h_l)$ is a binary indicator for VAF to indicate whether the visual content of the video clip where $h_l$ is extracted from is related to  its sound. $\mathbbm{1}_{VAF}(h_l)$ is set to $0$ only for inter modal contrastive loss when the CLIP similarity of visual frames and audio categories is smaller than a threshold.

Then the final loss of SVG-VQGAN is calculated as:
\begin{equation}
\begin{aligned}
\mathcal{L_{\text{SVG-VQGAN}}}=\mathcal{L}_{VQGAN}^v+\mathcal{L}_{VQGAN}^a+\alpha \mathcal{L}_{HCL}
\end{aligned}
\end{equation}

\subsection{Auto-Regressive Transformer Decoder}
\label{sec:Transformer}
As mentioned above, the text is tokenized by BPE \cite{bpe} as $X^{T}$=\{$x^t_1$,...,
$x^t_m$\}. The visual frames and audio signals are quantized into discrete tokens by the proposed SVG-VQGAN. In this section, we introduce the auto-regressive Transformer decoder to generate the visual and audio tokens with text tokens as input.
We utilize the unidirectional Transformer from Cogview \cite{cogview} as the backbone, and multimodal sequence formats are introduced for this sounding video generation task. Some specific tokens are used to indicate the modality or frame boundary. Specifically, we use $[TXT], [BOVi], [BOAi]$ to denote the beginning of text, the $i$-th visual frame and the $i$-th audio frame, respectively. $[EOVi]$ and $[EOAi]$ denote the end of the $i$-th visual frame and the $i$-th audio frame, respectively. Then modality cascade sequence format and modality alternate sequence format are introduced to build the input of auto-regressive Transformer decoder.

Modality cascade sequence format concatenates visual tokens $X^{V}$ and audio tokens $X^{A}$ as
\begin{equation}
\begin{aligned}
    X^{V}=\{[BOV1],x^{v}_1,[EOV1],...,[BOVL],x^{v}_L,[EOVL]\},\\
    X^{A}=\{[BOA1],x^{a}_1,[EOA1],...,[BOAL],x^{a}_L,[EOAL]\},
\end{aligned}
\end{equation}
where $x^{v}_i$ and $x^{a}_i$ denote the flattened discrete tokens of the $i$-th visual frame and the $i$-th audio frame. 
Then, all tokens are cascaded in the order of $[X^{T}, X^{V}, X^{A}]$ (T-V-A) or $[X^{T}, X^{A}, X^{V}]$ (T-A-V). Due to the unidirectional attention in auto-regressive Transformer, only visual-to-audio cross-modal association is built for T-V-A format and only audio-to-visual cross-modal association is built in T-A-V format.

\textbf{M}odality \textbf{A}lternate \textbf{S}equence \textbf{F}ormat (MASF) first concatenates both visual and audio tokens in a frame as
\begin{equation}
    X^{F}_i=\{[BOVi],x^{v}_i,[EOVi],[BOAi],x^{a}_i,[EOAi]\},
\end{equation}
 and then concatenates all frames with text token as
 \begin{equation}
     X=\{[TXT],X^T,X^F_1,...,X^F_L\}
 \end{equation}
In this way, the first visual frame is generated as a pivot and latter tokens could attend to both visual and audio content. 

The training object of auto-regressive Transformer decoder is left-to-right token prediction, using cross-entropy loss. 
All text, visual and audio tokens are equally treated, with  different loss weights $\gamma^v,\gamma^a,\gamma^t$,  
following Cogview \cite{cogview}. Finally the auto-regressive loss $\mathcal{L}_{AR}$ could be define as:
\begin{equation}
\begin{aligned}
\label{eq:lmloss}
    \mathcal{L}_{AR}^t&=-\gamma^t\sum_{i=1}^M\mathbbm{1}_{t}(X_i) X_i\log(P(X_i|X_{<i}))\\
    \mathcal{L}_{AR}^v&=-\gamma^v\sum_{i=1}^M\mathbbm{1}_{v}(X_i) X_i\log(P(X_i|X_{<i}))\\
    \mathcal{L}_{AR}^a&=-\gamma^a\sum_{i=1}^M\mathbbm{1}_{a}(X_i) X_i\log(P(X_i|X_{<i}))\\
    \mathcal{L}_{AR}&=\frac{\mathcal{L}_{AR}^t+\mathcal{L}_{AR}^v+\mathcal{L}_{AR}^a}{\sum_{i=1}^M\mathbbm{1}_{t}(X_i)\gamma^t+\mathbbm{1}_{v}(X_i)\gamma^v+\mathbbm{1}_{a}(X_i)\gamma^a}
\end{aligned}
\end{equation}
where $M$ is the length of $X$, $\mathbbm{1}_{t}(X_i)$, $\mathbbm{1}_{v}(X_i)$ and $\mathbbm{1}_{a}(X_i)$ separately denote whether $X_i$ is text, visual or audio tokens.
\section{Experiments}
\subsection{Datasets}
To solve the problem of missing appropriate training data for T2SV task, we construct a text-video-audio dataset based on AudioSet\cite{audioset}, named AudioSet-Cap. 
Audioset is an excellent data source as it is rich in audio diversity and provides links to the original videos.
Thus, we build the T2SV dataset by further supplementing the manually annotated text description for videos from AudioSet.
The annotators are required to describe both the visual and audio content, and filter out the low-quality data meeting the following conditions: 1) videos with meaningless visual or audio content which are hard to be described; 2) videos with no change through all frames; 3) videos less than 10 seconds (to train SVG with sufficiently long videos).
Finally there are 809,438 and 1,000 video clips of about 10 seconds each in the training set and test set. 
As shown in Table \ref{tab:dataset}, compared with other text-audio dataset, {such} as AudioCaps \cite{audiocaps}, and text-video dataset, such as HowTo-100M \cite{howto} and WebVid-2M \cite{webvid}, AudioSet-Cap contains meaningful audio signals in each video and provides accurate human annotated descriptions for both visual and background audio content.
These advantages make it more suitable for T2SV task. 
The dataset will be released soon.

To compare our method with state-of-the-art text-to-video and audio generation methods, we further evaluate our model on Kinetics \cite{kinetics} dataset as in T2V \cite{t2v}, and VAS \cite{vas} dataset as in SpecVQGAN \cite{specvqgan}. 
For the Kinetics dataset, we collect videos of the 10 classes first used in T2V\cite{t2v} from the original Kinetics\cite{kinetics} dataset and scrape there titles from the internet as the text descriptions.
Finally 5,186 video clips are selected as the training set and 1,000 videos from the original test set and part of the validation set are selected as the test set, as not all descriptions are available now due to invalid website and privacy. 
The VAS\cite{vas} dataset contains 9,520 and 754 video clips of 10 classes for training and evaluation. Videos less than 10 seconds are repeat and crop to keep the width of mel-spectrogram larger than 800, as in SpecVQGAN\cite{specvqgan}.

\begin{table}
\centering
  \caption{Comparison between AudioSet-Cap and other text-audio and text-video paired datasets. Note that the content and description respectively indicates whether there is only audio (A), only visual (v) or both (v+A) in the data and descriptions.}
  \label{tab:dataset}
  \begin{tabular}{ccccc}
    \toprule
    Dataset & \# clips & text source & content & description \\
    \midrule
    AudioCaps\cite{audiocaps} & 46k & Human & A & A \\
    \midrule
    HowTo-100M\cite{howto} & 136M & Internet & V+A & V \\
    WebVid-2M\cite{webvid} & 2.5M & Internet & V & V \\
    \midrule
    AudioSet-Cap & 0.8M & Human & V+A & V+A \\
  \bottomrule
\end{tabular}
\end{table}

\subsection{Implementation Details}
{ The raw audio with sampling rate of 22050 Hz is pre-processed as in SpecVQGAN \cite{specvqgan}, where a log-mel-spectrogram $\textbf{m}$ of size $(F\times T)=(80\times 800)$ is obtained, corresponding to a video clip of 9.26 seconds. 
The number of video sub-clips is $L=10$ and the size of the visual frames is set to $H=W=128$, which is a tradeoff between efficiency and video quality.
Thus, the visual frames are sparsely extracted from the raw video with $FPS=1/9.26=1.08$.} 
The dowonsampling rate $ds_v$ and $ds_a$ are both set to 16, resulting in visual frame tokens of size $(8\times 8)$ and audio frame tokens of size $(f\times t)=(5\times 5)$. 

The  encoders and  decoders follow the settings in VQGAN \cite{taming}, which are composed of convolutional stacks with skip-connections and group normalization.
There are 4 downsampling blocks in the encoders of SVG-VQGAN.
In each downsampling block, 2D-convolutional layers first spatially downsample the input by a factor of 2, then 2 residual blocks are used for feature extraction.
And the decoders are symmetric to the encoders, where convolution layers and nearest neighbor interpolation make up the upsampling blocks. 
We also add 2 self-attention layers at the end of the encoders, following VQGAN\cite{taming} and SpecVQGAN\cite{specvqgan}.
The output features of self-attention layers are considered as the output of encoders, which are then used for modeling cross-modal associations by CAM and quantized by Exponential Moving Average (EMA) vector quantizer. 
The dimensions $d_a, d_v$ are both set to 256.  The visual codebook size is $8192$ and the audio codebook size is $4096$. 
We use the modality split HCL according to the experiment. The similarity thresholds in VAF and TNS are set to 20.0 and 0.85 based on the statistics of dataset. And the window size in WPS is set to 2 for a larger batch size of 20, as we random crop 2 frames for each sampled video clip. After training for 700k iterations, we finetune SVG-VQGAN on 10 frames video clip and keep the window size of 2 with a batch size of 4 for 50k iterations to get better reconstruction quality of audio. The loss weight of HCL is set to 1.0. We optimize SVG-VQGAN using Adam \cite{adam} with a learning rate of 4.5e-6, on 4 NVIDIA-A100 GPUs.

The auto-regressive Transformer decoder is composed of 24 Transformer layers with 16-heads, and the dimension of hidden state is set to 1024. The max length of input sequence is set to 1025. We set the loss weight  according to the token length of different modalities, as $\gamma^t=3,\gamma^a=2,\gamma^v=1$. 
The parameters are updated by Adam \cite{adam} with a max learning rate of 8e-4. Warming up and cosine annealing decay \cite{sgdr} for learning rate are used. 
We train the Transformer on 8 NVIDIA-A100 GPUs with a batch size of 256 for 350k iterations.

The HiFiGAN\cite{hifigan} model is trained on AudioSet-Cap dataset with a batch size of 32, learning rate of 2e-4 and segment size of 8192 for 425k iterations. 
\subsection{Evaluation on Audioset-Cap}
\begin{table}
\centering
  \caption{Qualitative evaluation on AudioSet-Cap dataset. {K represents how many video samples are generated for a input text. T is the sampling time and sec denotes seconds. CLIPs represents CLIPSIM. * indicates whether use CLIPSIM to re-rank and select the generated videos matching best to the text.}}
  \label{tab:audioset}
  \begin{tabular}{cccccc}
    \toprule
    Method-K & T(sec) & CLIPs$\uparrow$ & FID-img$\downarrow$ & FID-vid$\downarrow$ & FID-aud$\downarrow$ \\
    \midrule
    { CogVideo-1} & {\textbf{276.96}} & {26.03} & {49.32} & {7.88} & {-} \\
    {CogVideo-4} & {412.58} & {26.00} & {\textbf{37.84}} & {7.40} & {-} \\
    {CogVideo*} & {412.58} & {\textbf{28.02}} & {48.09} & {\textbf{7.10}} & {-} \\
    \hline
    SVG-1 & {\textbf{39.77}} & 23.95 & 50.39 & 9.69 & 10.81 \\
    SVG-32 & {89.96} & 23.88 & \textbf{42.06} & 9.72 & \textbf{9.17} \\
    SVG* & {89.96} & \textbf{27.44} & 43.54 & \textbf{8.10} & 10.93 \\
  \bottomrule
\end{tabular}
\end{table}

\paragraph{\textbf{Quantitative Evaluation}}
We use the CLIPSIM metric proposed in GODIVA \cite{godiva} to measure the semantic consistency between text and video, which utilizes CLIP \cite{clip} to calculate the cosine similarity between the text and the generated visual frames. 
We further use FID-img \cite{fid} and FID-vid \cite{tfgan,3dresnet50} to evaluate the quality of generated visual frames as in TFGAN \cite{tfgan}. 
And FID-aud in SpecVQGAN \cite{specvqgan} is used for quantitative evaluation for the generated audio.  
To evaluate the semantic consistency between text-audio and visual-audio, manual evaluation is used. 
Manual evaluation score ranges from 0 to 100, where scores in (0, 25), [25,50), [50,75) and [75,100) indicate meaningless audio signals, audio signals mismatched with text, audio signals associated with text but not corresponding to video, 
audio signals matched with text and visual frames, respectively. 
{The evaluation criteria is the subjective evaluation of semantic consistency. 
If there is content that cannot be recognized semantically, it will be directly evaluated as semantically irrelevant. For example, if the audio signal matched the text description but the visual fidelity is too low, it will be evaluated to [50,75).}
32 samples are generated for each text description and CLIPSIM \cite{clip,godiva} is used to find the video matching best to the text.

\begin{figure}
  \centering
  \includegraphics[width=0.9\linewidth]{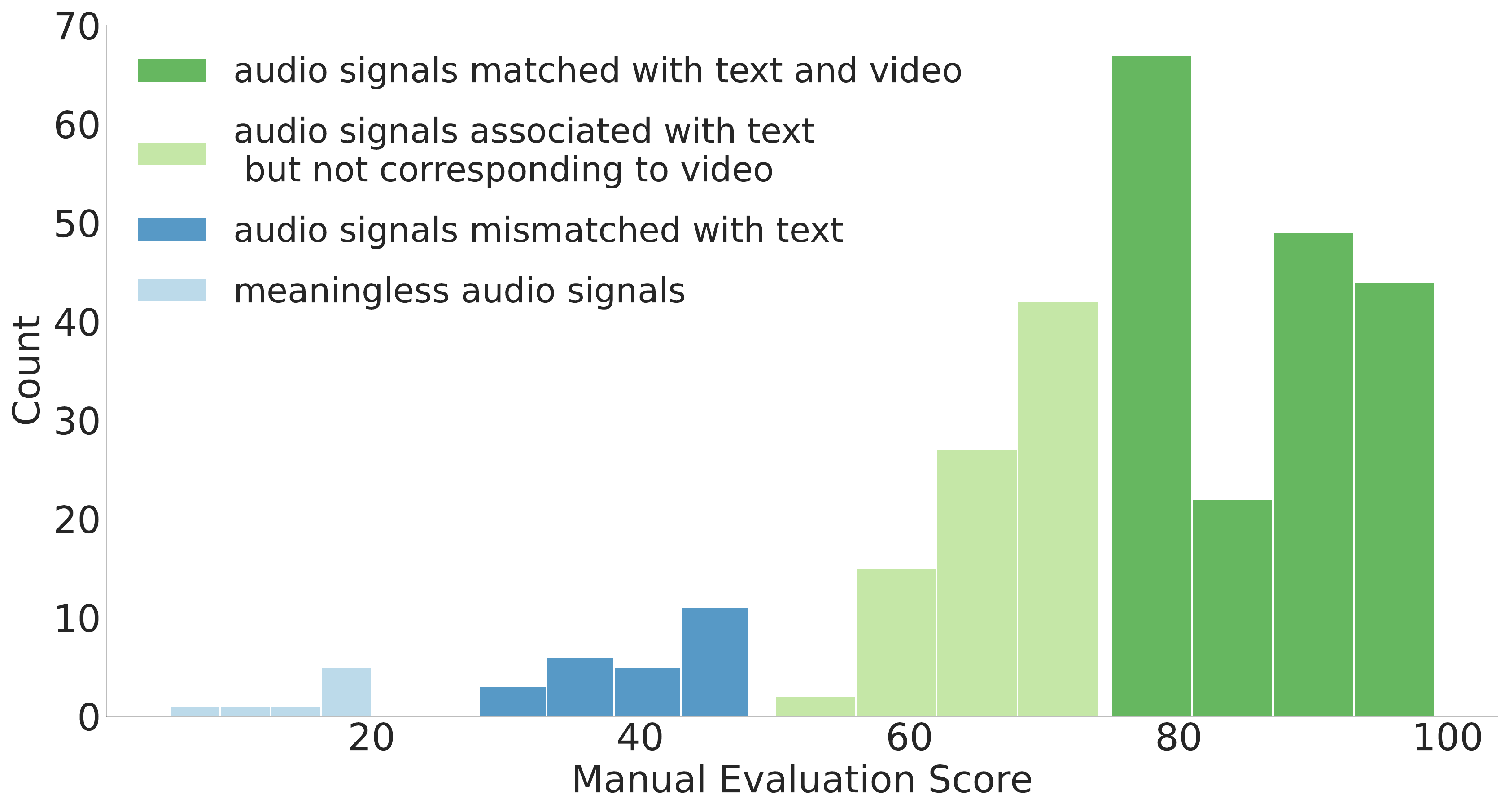}
  \caption{Histogram of manual evaluation scores of 300 videos generated by SVG and re-ranked using CLIPSIM, guided by texts randomly sampled from the AudioSet-Cap test set.}
  \label{fig:ME}
  \vspace{-4mm}
\end{figure}

As shown in Table \ref{tab:audioset}, better semantic consistency is obtained after re-ranked by CLIPSIM, along with better realism of generated visual frames for the smallest FID-vid. Using all samples gets better FID-img and FID-aud because more samples are more likely to fit the distribution of real visual frames and audio. 
{We compare our method with state-of-the-art two-stage video generation model, i.e., CogVideo \cite{cogvideo}.
5 frames are generated for each video as in CogVideo-stage1. 
To prevent the difference caused by frame number, we repeat the generated video to 10 frames to calculate FID-img and FID-vid.
It could be found that CogVideo generates video frames with higher fidelity and better visual-text consistency, 
as more parameters and frames with higher resolution of $480\times480$ are used by CogVideo, which also leads to slower inference.
Besides, SVG could generate associated audio while CogVideo focuses on video frames generation.}
Fig. \ref{fig:ME} shows the manual evaluation result of videos generated by SVG with CLIPSIM re-rank. It can be found that most of the audio signals, visual frames and texts are associated.
\begin{figure*}
  \centering
  \includegraphics[width=0.95\linewidth]{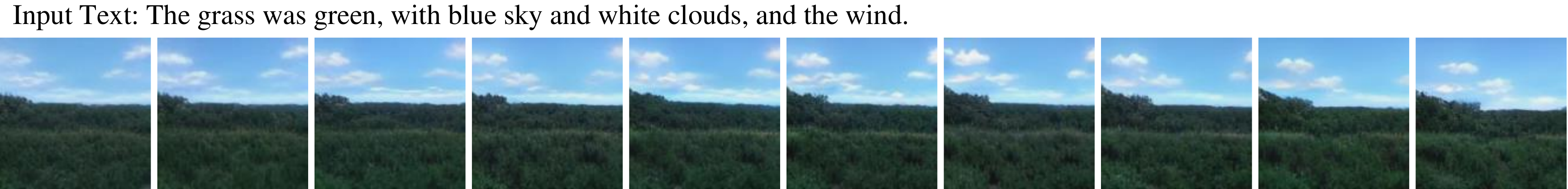}
  \includemedia[
  addresource=media/wind.wav,
  transparent,
  flashvars={
    source=media/wind.wav
  &autoPlay=true
  },
]{\includegraphics[width=0.95\linewidth]{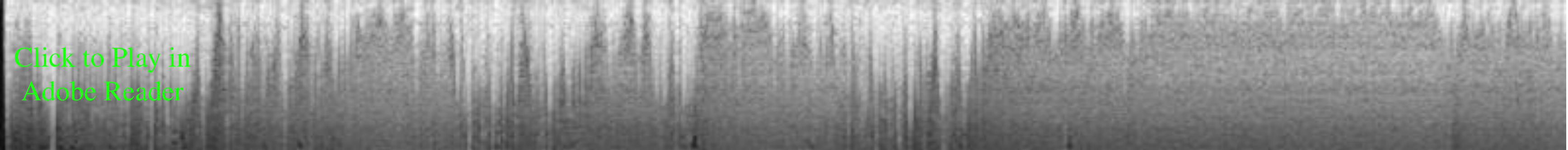}}{APlayer.swf}
\includegraphics[width=0.95\linewidth]{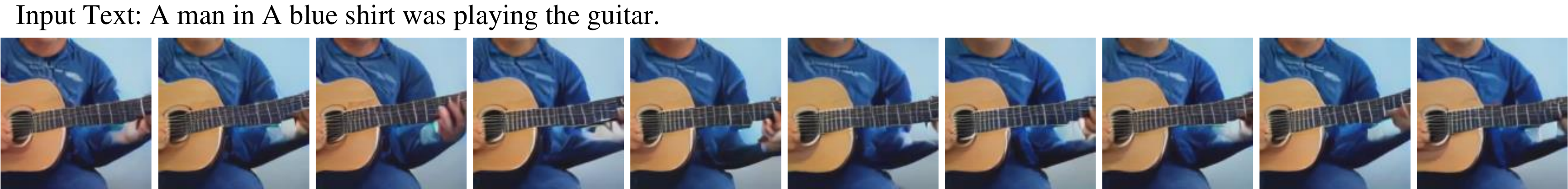}
  \includemedia[
  addresource=media/guitar.wav,
  transparent,
  flashvars={
    source=media/guitar.wav
  &autoPlay=true
  },
]{\includegraphics[width=0.95\linewidth]{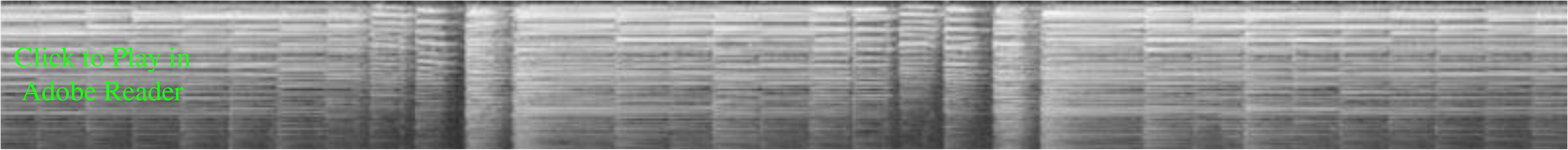}}{APlayer.swf}
\includegraphics[width=0.95\linewidth]{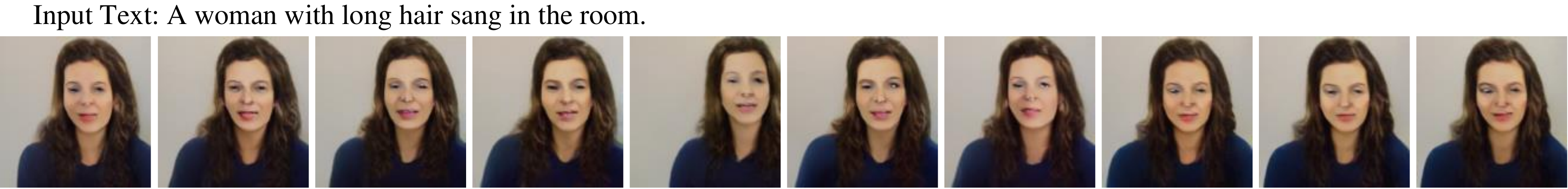}
  \includemedia[
  addresource=media/woman.wav,
  transparent,
  flashvars={
    source=media/woman.wav
  &autoPlay=true
  },
]{\includegraphics[width=0.95\linewidth]{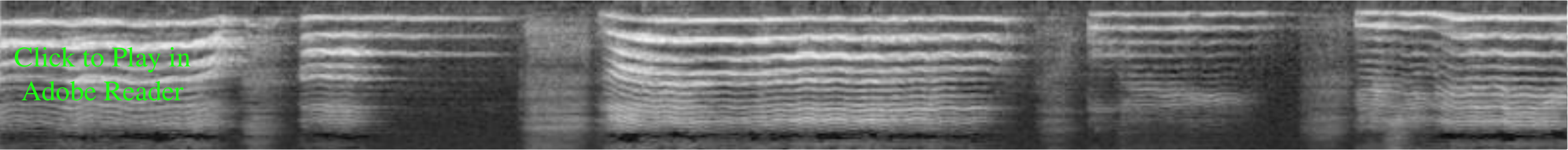}}{APlayer.swf}
\includegraphics[width=0.95\linewidth]{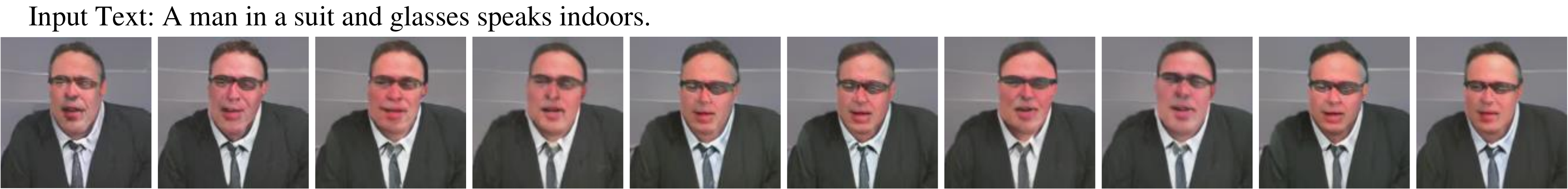}
  \includemedia[
  addresource=media/man.wav,
  transparent,
  flashvars={
    source=media/man.wav
  &autoPlay=true
  },
]{\includegraphics[width=0.95\linewidth]{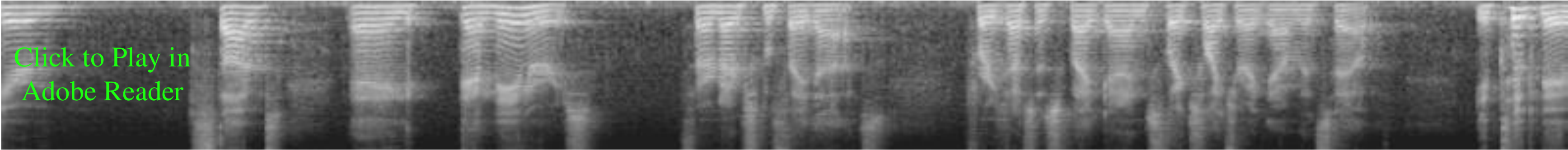}}{APlayer.swf}
\includegraphics[width=0.95\linewidth]{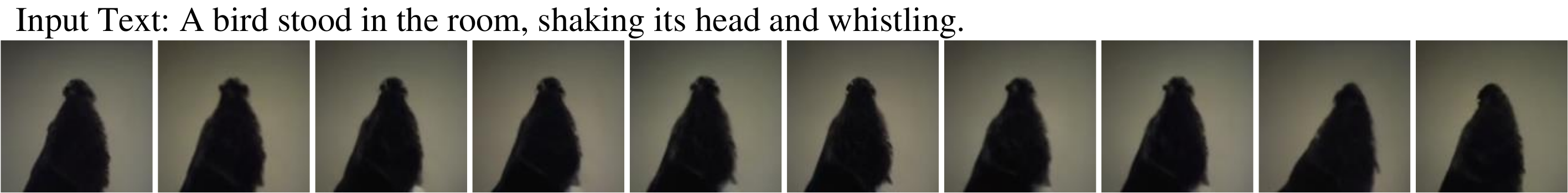}
  \includemedia[
  addresource=media/bird.wav,
  transparent,
  flashvars={
    source=media/bird.wav
  &autoPlay=true
  },
]{\includegraphics[width=0.95\linewidth]{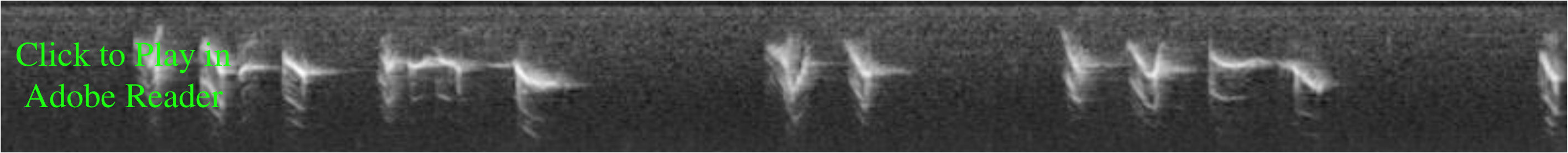}}{APlayer.swf}
\includegraphics[width=0.95\linewidth]{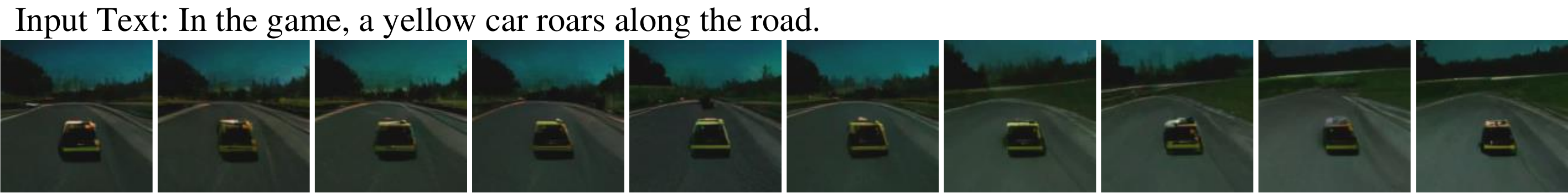}
  \includemedia[
  addresource=media/car.wav,
  transparent,
  flashvars={
    source=media/car.wav
  &autoPlay=true
  },
]{\includegraphics[width=0.95\linewidth]{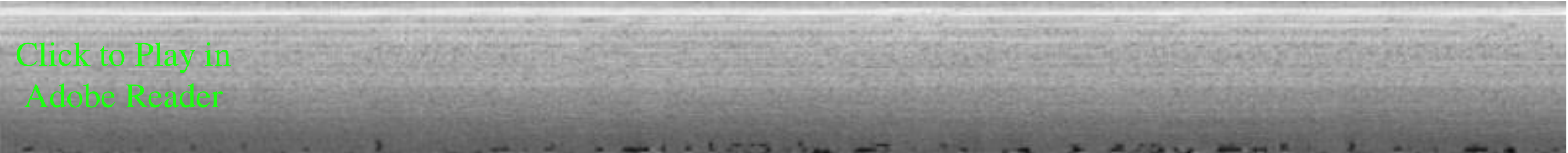}}{APlayer.swf}
  \caption{{Visualization of various generated visual frames and mel-spectrograms, containing landscapes, animals, objects, and humanities.} The columns of mel-spectrograms from top to bottom represent low to high frequencies and the rows represent changes over time. Audio files could also be found in \url{https://github.com/jwliu-cc/SVG.git} when Adobe Reader is unavailable.}
  \label{fig:demo}
\end{figure*}

\paragraph{\textbf{Qualitative Evaluation}}
Visualization of {various} generated frames and mel-spectrograms are shown in Fig. \ref{fig:demo}.
It can be found that the visual frames generated by SVG match the text description well, and the generated audio signals also present the sound characteristics, e.g., the wind is concentrated on the low frequency, the sound of guitar is rhythmic, and the frequency of the human voice is richer.
On the other hand, thanks to the modality alternate sequence format, the audio tokens can only attend to the previous visual frames, so that the generated audio and visual frames have a certain time alignment, such as the example of a woman singing.
More synthesised videos could be found in the project page
\footnote{\url{https://github.com/jwliu-cc/SVG.git}}. 

\subsection{Evaluation on Open-Sourced Dataset}
In this section, we compare our method with state-of-the-art video generation methods on Kinetics \cite{kinetics,t2v} dataset and audio generation method on VAS \cite{vas} dataset, while we generate both visual frames and audio signals guided by text descriptions simultaneously. 

\begin{table}
\centering
  \caption{Performance comparison with text-to-video generation methods on Kinetics dataset.}
  \label{tab:kinetics}
  \begin{tabular}{cccc}
    \toprule
    Model & CLIPSIM $\uparrow$ & FID-img $\downarrow$ & FID-vid $\downarrow$ \\
    \midrule
    T2V ($64\times64$) \cite{t2v} & 28.53 & 82.13 & 14.65  \\
    SC ($128\times128$) \cite{tfgan} & 29.15 & 33.51 & 7.34  \\
    TFGAN ($128\times128$) \cite{tfgan} & 29.61 & 31.76 & 7.19  \\
    N{\"U}WA ($128\times128$) \cite{nuwa} & \textbf{30.12} & 28.46 & 7.05  \\
    \midrule
    SVG ($128\times128$) & 29.72 & \textbf{27.45} & \textbf{5.19}  \\
  \bottomrule
\end{tabular}
\end{table}

\begin{small}
\begin{table}
\centering
  \caption{Performance comparison of text-to-audio generation on VAS dataset. $\dag$ means using visual frames as extra input.}
  \label{tab:vas}
  \begin{tabular}{cccc}
    \toprule
    Model & Training set & FID-aud $\downarrow$ & MKL-aud $\downarrow$\\
    \midrule
    \multirow{2}*{SpecVQGAN \cite{specvqgan}} & VGGSound \cite{vggsound} & 33.7 & 9.6 \\
    & VAS \cite{vas} & 28.7 & 9.2 \\
    \midrule
    \multirow{2}*{$\text{SpecVQGAN}^\dag$ \cite{specvqgan}} & VGGSound \cite{vggsound} & 20.5 & 6.0  \\
    & VAS \cite{vas} & 22.6 & 5.8 \\
    \midrule
    \multirow{2}*{SVG (ours)} & Audioset-Cap & 39.03  & 9.66 \\
     & VAS \cite{vas} & \textbf{9.00} & \textbf{4.67} \\
  \bottomrule
\end{tabular}
\end{table}
\end{small}
\begin{figure*}
  \centering
  \includegraphics[width=0.95\linewidth]{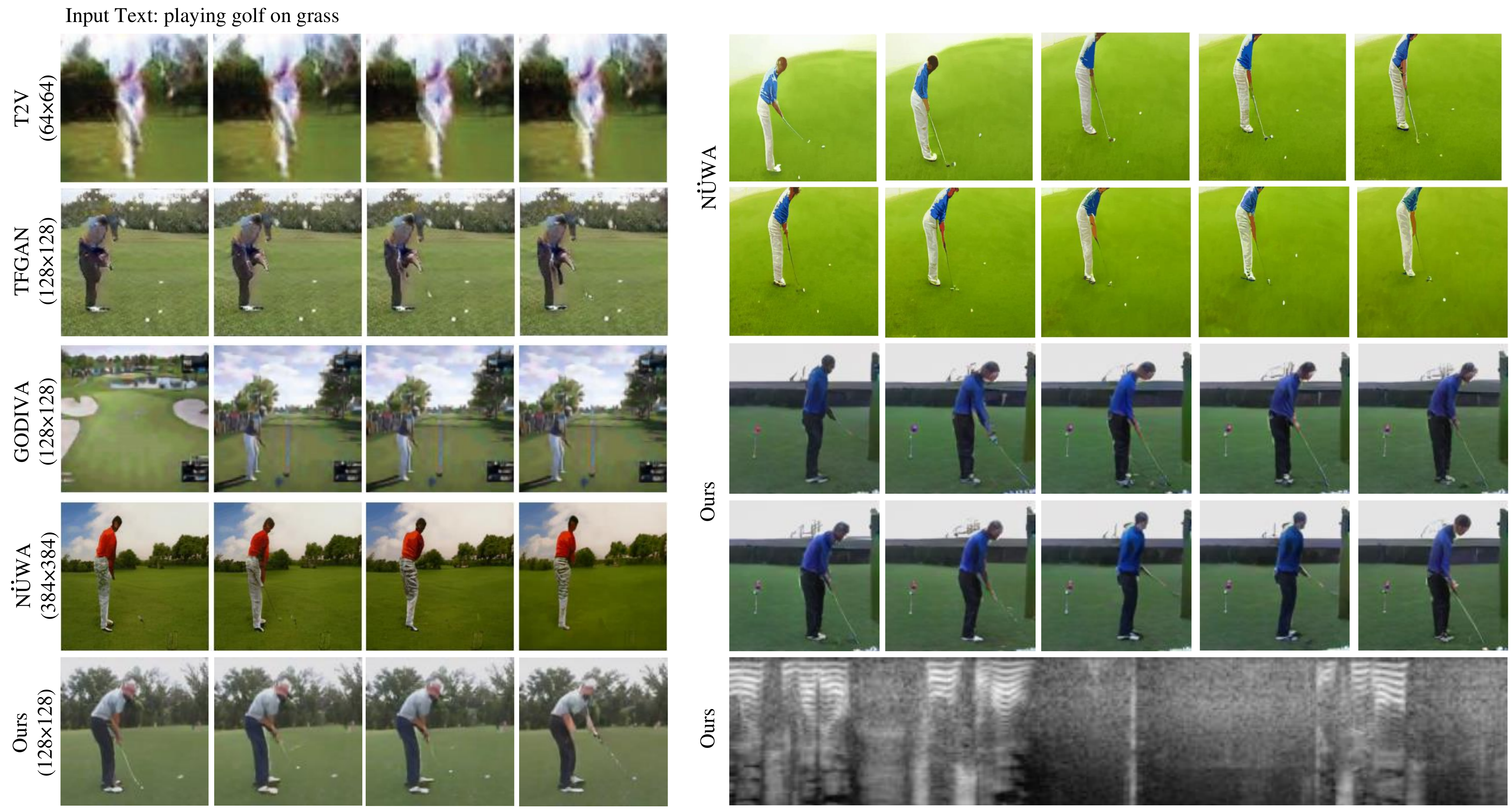}
  \caption{Visualization of text-to-video generation on Kinetics dataset.}
  \label{fig:kinetics}
\end{figure*}
\begin{figure}
  \centering
  \includegraphics[width=0.95\linewidth]{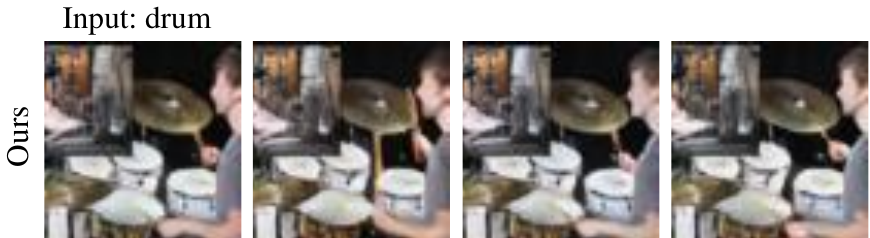}
  \includemedia[
  addresource=media/drumGT.wav,
  transparent,
  flashvars={
    source=media/drumGT.wav
  &autoPlay=true
  },
]{\includegraphics[width=0.95\linewidth]{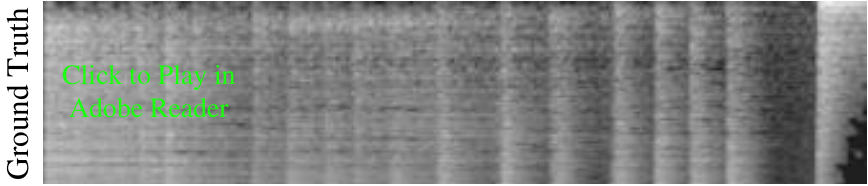}}{APlayer.swf}
\includemedia[
  addresource=media/drumSVG.wav,
  transparent,
  flashvars={
    source=media/drumSVG.wav
  &autoPlay=true
  },
]{\includegraphics[width=0.95\linewidth]{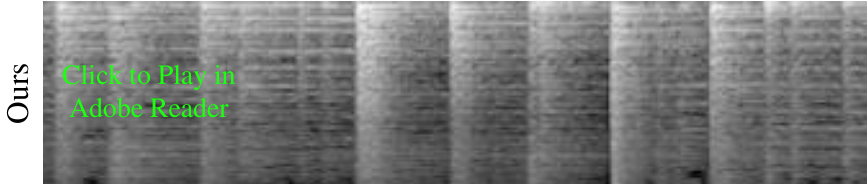}}{APlayer.swf}
\includemedia[
  addresource=media/drumSpecVQGAN.wav,
  transparent,
  flashvars={
    source=media/drumSpecVQGAN.wav
  &autoPlay=true
  },
]{\includegraphics[width=0.95\linewidth]{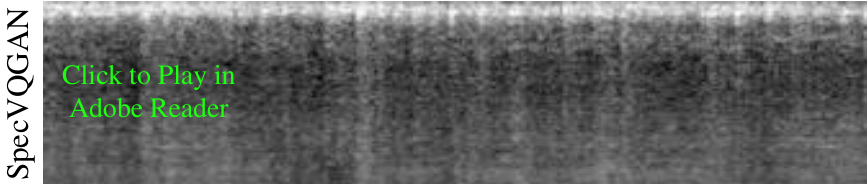}}{APlayer.swf}
  \caption{Visualization of text-to-audio generation on VAS dataset. Note that the visual frames  in the first line are generated by our model while SpecVQGAN takes real visual frames as input.}
  \label{fig:vas}
\end{figure}

\paragraph{\textbf{Text-to-Video Generation}}
We compare our method with other text-to-video generation methods quantitatively in Table \ref{tab:kinetics} and qualitatively in Fig. \ref{fig:kinetics}.
Note that we separately finetune SVG-VQGAN and Transformer on Kinetics for 10 epochs and 2k iterations, then generate 32 samples for each text and re-rank with CLIPSIM. As shown in Table \ref{tab:kinetics}, our proposed SVG outperforms previous text-to-video generation methods in most metrics, and is comparable to N{\"U}WA \cite{nuwa} in CLIPSIM. 
Visualization of generated samples could be found in Fig. \ref{fig:kinetics}.
It can be found that the quality of the visual frames generated by our model is better than previous generation models T2V \cite{t2v}, TFGAN \cite{tfgan}, GODIVA \cite{godiva},  while we could generate audio signals at the same time. Note that N{\"U}WA \cite{nuwa} generates video with a high resolution of $384\times384$, leading to better visualization but longer visual tokens sequence, which also introduces greater computational consumption.

\paragraph{\textbf{Text-to-Audio Generation}}
We set the $N^v=512$ and $N^a=128$ when training on VAS for fair comparison with SpecVQGAN \cite{specvqgan} and the resolution of visual frames is $64\times 64$ with downsampling rate 8 for this small dataset. The class labels in VAS are taken as the input text. We use the FID-aud and MKL metric for quantitative evaluation as in SpecVQGAN \cite{specvqgan}. Results in Table \ref{tab:vas} show that our method trained on VAS remarkably outperforms SpecVQGAN even when SpecVQGAN uses visual frames as extra input. We also prove the zero-shot generation result of our model when trained on AudioSet-Cap dataset. Since the model is trained using description as input, the performance is slightly inferior to SpecVQGAN trained on VGGSound \cite{vggsound}, which using class names as input as VAS. As shown in Fig. \ref{fig:vas}, The mel-spectrograms generated by our model is smoother and clearer than SpecVQGAN. Note that the generated audio of SpecVQGAN is downloaded from the project page\footnote{\url{https://iashin.ai/SpecVQGAN}}.  

\begin{table}
  \centering
  \caption{Ablation study of SVG-VQGAN. All experiments are trained with HCL except for \expandafter{\romannumeral1}-2. * denotes audio categories in VAF are extracted by pretrained PaSST\cite{passt} otherwise from the labels in AudioSet. $\dag$ represents finetuning with 10 frames and original window size for further 20k iterations.}
  \label{tab:SVG-VQGAN}
  \begin{tabular}{c|cccc|cc}
    \toprule
     & MS & VAF & TNS & WPS & FID-aud $\downarrow$ & FID-img $\downarrow$ \\
    \midrule
    \expandafter{\romannumeral1}-1 & \cmark & 20.0 & 0.85 &2 & 10.20 & \textbf{22.14} \\
    \expandafter{\romannumeral1}-2 & - & - & - & 2 & 10.87 (+0.67) & 23.30 (+1.16) \\
    \midrule
    \expandafter{\romannumeral2}-1 & \cmark & 20.0* & 0.85 &2 & 10.39 (+0.19) & 22.36 (+0.22) \\
    \expandafter{\romannumeral2}-2 & \xmark & 20.0 & 0.85 & 2 & 10.83 (+0.63) & 23.24 (+1.10) \\
    \expandafter{\romannumeral2}-3 & \cmark & - & 0.85 & 2 & 10.24 (+0.04) & 23.61 (+1.47) \\
    \expandafter{\romannumeral2}-4 & \cmark & 20.0 & - & 2 & 11.03 (+0.83) & 23.15 (+1.01) \\
    \midrule
    {\expandafter{\romannumeral3}-1} & {\cmark} & {22.0} & {0.85} & {2} & {9.18 (-1.02)} & {21.74 (-0.40)} \\
    {\expandafter{\romannumeral3}-2} & {\cmark} & {18.0} & {0.85} & {2} & {10.14 (-0.06)} & {23.05 (+0.91)} \\
    {\expandafter{\romannumeral3}-3} & {\cmark} & {20.0} & {0.80} & {2} & {10.33 (+0.13)} & {22.59 (+0.45)} \\
    {\expandafter{\romannumeral3}-4} & {\cmark} & {20.0} & {0.90} & {2} & {10.44 (+0.24)} & {22.66 (+0.52)} \\
    \expandafter{\romannumeral3}-5 & - & 20.0 & 0.85 & 1 & 10.18 (-0.02) & 23.43 (+1.29)\\
    \expandafter{\romannumeral3}-6 & \cmark & 20.0 & 0.85 & 4 & 7.67 (-2.53) & 22.81 (+0.67) \\
    \expandafter{\romannumeral3}-7 & \cmark & 20.0 & 0.85 & 10 & \textbf{7.58} (-2.62) & 27.63 (+5.49) \\
    \midrule
    \expandafter{\romannumeral4}-1 & \cmark & 20.0 & 0.85& $4^\dag$ & \textbf{7.01} (-3.19) & 21.13 (-1.01)\\
    \expandafter{\romannumeral4}-2 & \cmark & 20.0 & 0.85& $2^\dag$ & 7.11 (-3.09) & \textbf{20.31} (-1.83)\\
  
  \bottomrule
\end{tabular}
\end{table}
\begin{table}      
  \centering
  \caption{Ablation study for training Transformer decoder with SVG-VQGAN with and without HCL.}
  \label{tab:ab_HCL_trm}
  \begin{tabular}{ccccc}
    \toprule
     & CLIPSIM$\uparrow$ & FID-img$\downarrow$ & FID-vid$\downarrow$ & FID-aud$\downarrow$ \\
    \midrule
    w/o HCL & 26.27 & 78.79 & 16.15 & 25.06 \\
    w/ HCL & \textbf{26.45} & \textbf{76.72} & \textbf{15.08} & \textbf{22.12} \\
  \bottomrule
\end{tabular}
\end{table}
\begin{figure}
  \centering
  \includegraphics[width=\linewidth]{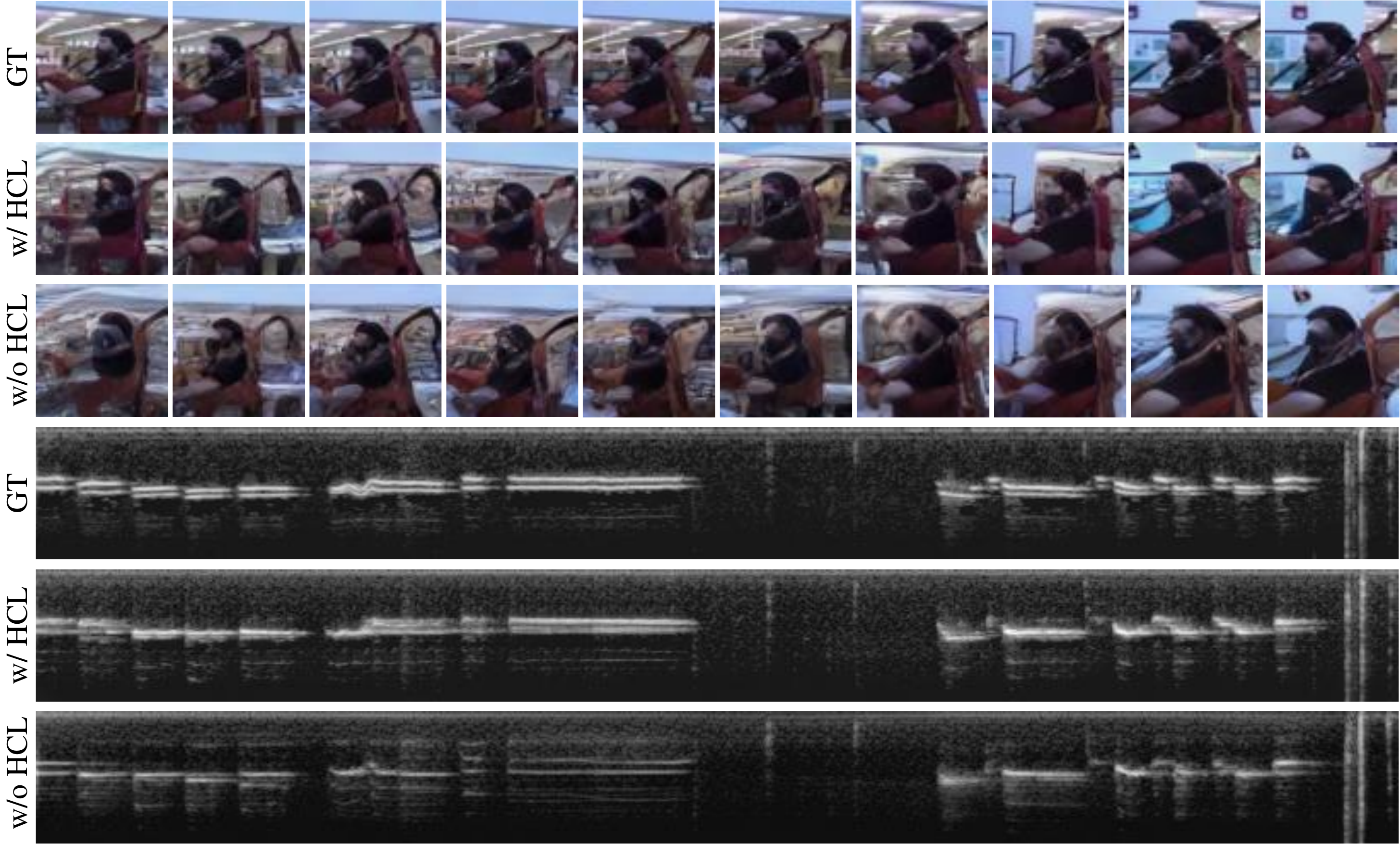}
  \caption{Visualization of visual frames and mel-spectrograms of ground truth (GT), reconstruction results of SVG-VQGAN with HCL (w/ HCL) and without HCL (w/o HCL). HCL makes SVG-VQGAN pay more attention to key feature areas and achieve better reconstruction quality.
  }
  \label{fig:recon1}
\end{figure}
\begin{figure}
  \centering
  \includegraphics[scale=0.198]{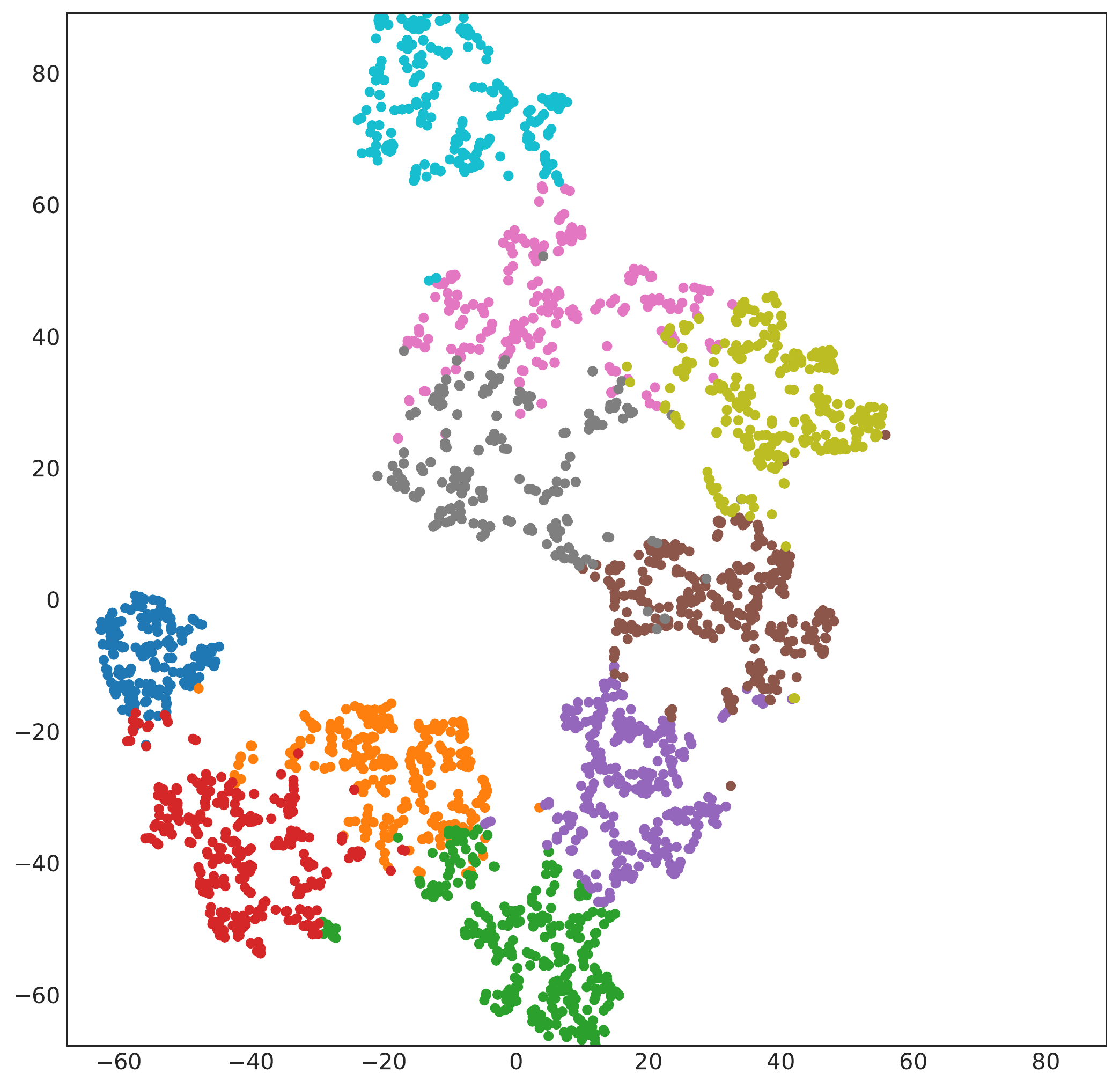}
  \includegraphics[scale=0.198]{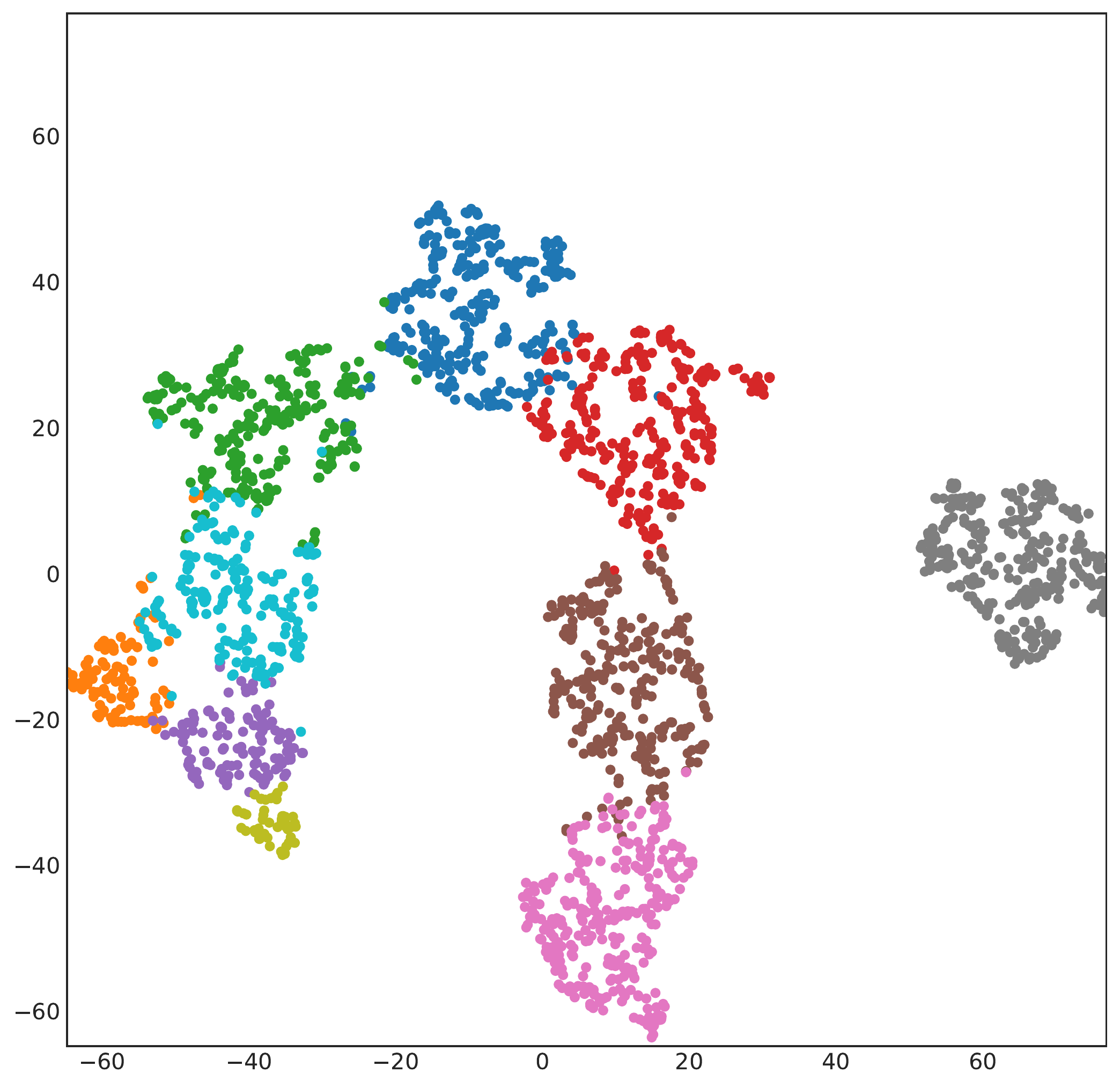}
  \caption{t-SNE visualization of visual features encoded by SVG-VQGAN without HCL (left) and with HCL (right). Different colors indicate different codebook embeddings.}
  \label{fig:tsne}
\end{figure}
\subsection{Ablation Study}
\paragraph{\textbf{SVG-VQGAN at Reconstruction Stage}}
For the ablation study at the reconstruction stage, we set the resolution of visual frames to $64\times 64$ with a downsampling rate of 8 for simplicity, and train on AudioSet-Cap for 10 epochs, with $N^v=N^a=2048$. 

Quantitative results could be found in Table \ref{tab:SVG-VQGAN}, {including experiments with or without HCL (\expandafter{\romannumeral1}), ablation study of different settings of HCL (\expandafter{\romannumeral2}), ablation study of hyper parameters (\expandafter{\romannumeral3}) and 10-frames finetuning experiments (\expandafter{\romannumeral4}).} 
Improvements have been achieved on both FID-aud and FID-img when training SVG-VQGAN with additionally HCL (comparing \expandafter{\romannumeral1}-1 and \expandafter{\romannumeral1}-2). 
It is worth noting that using HCL with accurate audio category annotations (\expandafter{\romannumeral1}-1) or categories extracted by pretrained PaSST \cite{passt} (\expandafter{\romannumeral2}-1) for VAF have closer performance, which shows that HCL is universal and can be used for other datasets without audio category annotations.

{We conducted experiments (\expandafter{\romannumeral2}) on replacing modality split contrastive loss with modality gathered contrastive loss, removing VAF and TNS strategies.} 
Modality split contrastive loss is shown to be better than modality gathered contrastive loss (\expandafter{\romannumeral2}-2) in this reconstruction task, as it separates the construction of cross-modal correlation and the regularity of of intra-modal distribution.
Removing VAF (\expandafter{\romannumeral2}-3) does harm to the performance of SVG-VQGAN, as video clips with uncorrelated visual-audio content are used as positive samples.
It should be noted that the reconstruction quality is also degraded when TNS is removed (\expandafter{\romannumeral2}-4), especially on audio,  because there is a large number of audios with similar semantics in AudioSet dataset, e.g., concert videos, and it is critical to use text descriptions for selecting semantically distinct negative samples.

{From the ablation study of hyper parameters, it could be found that SVG-VQGAN with different VAF (\expandafter{\romannumeral3}-1 and \expandafter{\romannumeral3}-2) and TNS (\expandafter{\romannumeral3}-3 and \expandafter{\romannumeral3}-4) thresholds outperforms SVG-VQGAN without HCL (\expandafter{\romannumeral1}-2), which shows the robustness of HCL. 
It is worth noting that improvements have been achieved in SVG-VQGAN with a higher VAF threshold of 22.0 (\expandafter{\romannumeral3}-1), as better inter-modal positive samples are provided.
However, we set the VAF threshold to 20.0 because accurate audio labels may be missing in other datasets.}
We also notice that the quality of video reconstruction decreases when the window size in WPS is set to 1 (\expandafter{\romannumeral3}-5), which shows the importance of intra-modal contrastive loss.
Using a larger window size (i.e. 4 and 10) in WPS (\expandafter{\romannumeral3}-6 and \expandafter{\romannumeral3}-7) is better for audio spectrogram reconstruction, as we evaluate on the whole audio mel-spectrogram of 10 frames and train the model with larger window size can better adapt to the 10-frames mel-spectrogram. 
However, the performance of video reconstruction is degraded because some positive samples with poor correlation may be introduced and the frame diversity within a batch is reduced. 
Besides, it is hard to train with large window size when using higher resolution visual frames, limited by the memory of GPUs. 
Actually, we can further finetune the  SVG-VQGAN with 10-frames video to achieve better quality in audio mel-spectrogram reconstruction, as shown in the last two rows of Table \ref{tab:SVG-VQGAN}.

Comparison of some visualized reconstructed examples are shown in Fig. \ref{fig:recon1}, and the obvious advantages of HCL can be found in the reconstruction of audio mel-spectrogram, where the areas with significant characteristics related to visual content will be particularly focused on, which will be discussed later, and reconstructed better.

We further visualise the visual features encoded by SVG-VQGAN with and without HCL in Fig.\ref{fig:tsne}. The encoded features in the validation set corresponding to the 10 embeddings with the highest cosine similarity in the video codebook are dimensionally reduced by t-SNE \cite{tsne} and visualized. 
It is obviously that the features extracted by SVG-VQGAN with HCL are more clustering and separable, while the gray, pink and yellow-green features extracted by SVG-VQGAN without HCL are dispersive and mixed.
\paragraph{\textbf{SVG-VQGAN on Generation Stage}}
For the ablation study of SVG-VQGAN with and without HCL on generation stage,  
we use the pretrained SVG-VQGAN with and without HCL above to extract visual and audio tokens, and construct Transformers with 12 self-attention layers and the hidden size of 1024 for auto-regressive token generation. Both of the Transformer models are trained on AudioSet-Cap dataset for 100k iterations. 
Results could be found in Table \ref{tab:ab_HCL_trm}. It can be found that the generation quality of SVG using SVG-VQGAN with HCL is better than that without HCL on all evaluation metrics, which indicates that HCL improves the quantized representations of visual frames and audio signals and benefits the training of Transformer because of less noise.

\begin{table}[t]
  \centering
  \caption{Ablation study on different sequence formats.}
  \label{tab:abgt}
  \begin{tabular}{cccc}
    \toprule
    Sequence Format & CLIPSIM $\uparrow$ & FID-img $\downarrow$ & FID-aud $\downarrow$ \\
    \midrule
    T-V-A & 26.19 & 69.50 & 16.69 \\
    T-A-V & 26.14 & 69.52 & 16.28 \\
    \midrule
    MASF & \textbf{26.33} & \textbf{66.31} & \textbf{16.08}  \\
  \bottomrule
\end{tabular}
\end{table}

\paragraph{\textbf{Different Multi-modal Sequence Formats}}
For the ablation study of different sequence formats, we use a smaller version Transformer with 12 layers and train the model for 100k iterations on AudioSet-Cap. 16 samples are generated for each text and all samples are used for calculating FID-img and FID-aud. \textbf{M}odality \textbf{A}lternate\textbf{S}equence \textbf{F}ormat (MASF) outperforms modality cascade sequence format, i.e. T-V-A and T-A-V, on all metrics. The reasons are from two aspects. On the one hand, MASF can build cross-modal associations in both audio-to-visual and visual-to-audio, while T-V-A and T-A-V only focus on single directional cross-modal associations. 
On the other hand, T-V-A and T-A-V are more dependent on the quality of the previous generated modality, making it susceptible to previous generation errors.

\begin{figure}
  \centering
  \includegraphics[scale=0.198]{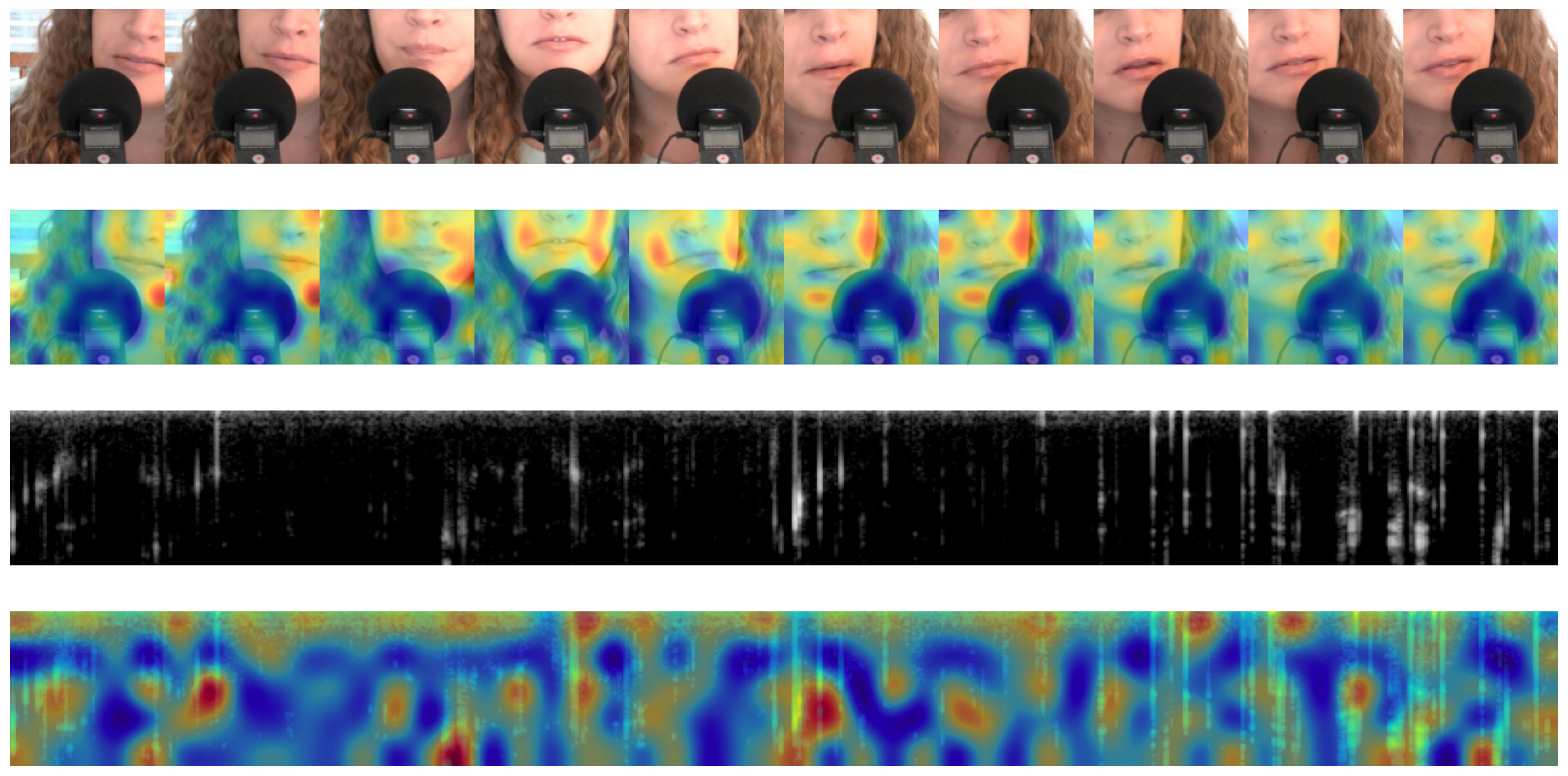}
  \caption{An example of attention map obtained by CAM.}
  \label{fig:att_examples}
\end{figure}

\subsection{Visualization of Cross-modal Attention Module}
We visualise the attention map in CAM in Fig. \ref{fig:att_examples}. For visual attention map, we take the average through all 5 audio frequency bands. The visualised example shows that the audio-to-visual attention could capture the main visual position where the sound comes from, e.g., the face of the woman in the example. 
And more attention was paid to the areas with prominent features in the audio mel-spectrogram in visual-to-audio attention.
Therefore, local alignment between visual frames and audio mel-spectrograms is build, through which audio-associated visual features and 
visual-associated audio features are obtained for HCL.

\section{Conclusion}
In this paper, we present Sounding Video Generator (SVG) as a unified model, that can simultaneously generate video with audio signals  guided by text descriptions for the first time. 
A novel SVG-VQGAN with cross-modal attention module and hybrid contrastive loss is proposed to quantize visual frames and audio mel-spectrograms into discrete tokens. 
Then an auto-regressive Transformer decoder with a modality alternate sequence format is used for generating visual and audio tokens guided by the text descriptions.
In this way, SVG could model visual-audio associations at both the encoding and decoding stage, and generate semantically associated visual frames and audio signals guided by text. 
Future studies may include high resolution and high frame rate video generation, and more explicit modeling of the temporal alignment between visual frames and audio signals.
\section{Acknowledgement}
This work was supported by the National Key Research and Development Program of China (No. 2020AAA0106400), National Natural Science Foundation of China (U21B2043, 62102419, 62102416) and CAAI-Huawei MindSpore Open Fund.
\bibliographystyle{IEEEtran}
\bibliography{main_ref}

\vfill

\end{document}